\documentclass{article}
\usepackage[makeroom]{cancel}
\usepackage[utf8]{inputenc}
\usepackage{microtype}
\usepackage{graphicx}
\usepackage{subfigure}

\usepackage[accepted]{icml2020}
\usepackage{mathtools}

\newcommand{\SFig}{Supplementary Fig}
\newcommand{\caffe}{$_{\text{caffe}}$}
\newcommand{\slim}{$_{\text{slim}}$}
\newcommand{\y}{\mathbf{f}}
\newcommand{\s}{\mathbf{s}}
\renewcommand{\a}{\mathbf{a}}
\newcommand{\ahl}{\a^{h, l}}
\newcommand{\ax}{a}
\newcommand{\abl}{\a^{b, l}}
\newcommand{\x}{{\mathbf{x}}}

\newcommand{\yb}{\y^{b=0}}
\newcommand{\abs}{\mathrm{abs}}

\newcommand{\vgg}{VGG-16}
\newcommand{\resnet}{ResNet-50}
\newcommand{\dxy}[2]{\frac{\partial{#2}}{\partial{#1}}}
\newcommand{\e}{e}
\newcommand{\ee}{\e_{\mathrm{-}}}
\newcommand{\bxi}{\mathbf{\xi}}
\newcommand{\bn}{BatchNorm}

\begin{document}

\twocolumn[
\icmltitle{Measuring and improving the quality of visual explanations}
\begin{icmlauthorlist}
\icmlauthor{Agnieszka Grabska-Barwińska}{}
\end{icmlauthorlist}
\icmlcorrespondingauthor{Agnieszka Grabska-Barwińska}{agnigb@google.com}
\icmlkeywords{interpretability, saliency, explanation}

\vskip 0.3in
]

\begin{abstract}
The ability of to explain neural network decisions goes hand in hand with their safe deployment. Several methods have been proposed to highlight features important for a given network decision. 
However, there is no consensus on how to measure effectiveness of these methods. 
We propose a new procedure for evaluating explanations. We use it to investigate visual explanations extracted from a range of possible sources in a neural network. We quantify the benefit of combining these sources and challenge a recent appeal for taking account of bias parameters. We support our conclusions with a general assessment of the impact of bias parameters in ImageNet classifiers.
\end{abstract}

\section{Introduction}\label{sec:introduction}
Neural networks are well established in a wide range of applications, such as classification, recommendation, natural language understanding, to name but a few. Their deployment however, is accompanied by reservation and an increasing demand for explanation \cite{GoodmanFlaxman2017}. Here, we add to the research effort focused on explaining single decisions---information that is most relevant in medical diagnosis, law, banking, and any other field where machine learning supports human decisions. Instance-based explanations attempt to delineate factors that are most important for the given network decision.

\begin{figure}
\subfigure[Gradients in a linear digit classifier]{
    \includegraphics[width=\columnwidth]{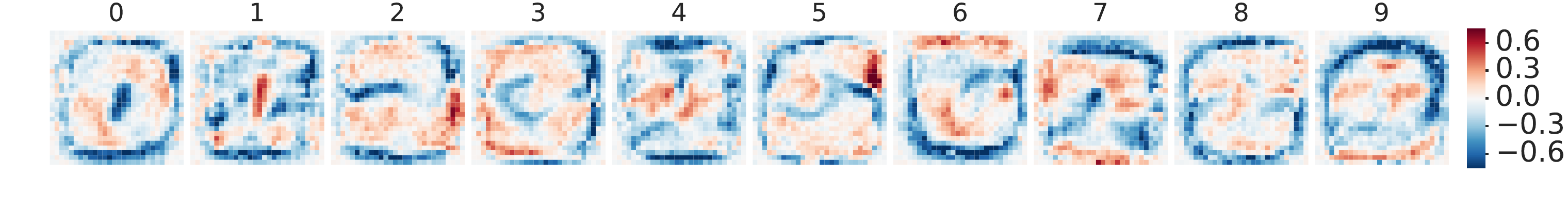}\label{fig:explanation_a}
    }
    \includegraphics[width=\columnwidth]{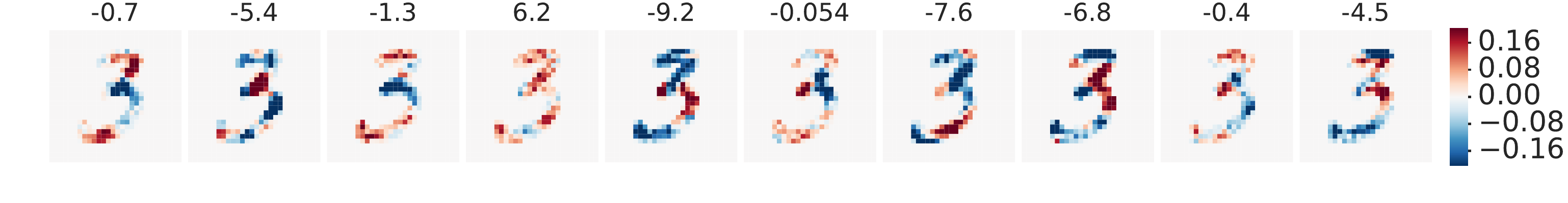}\\
\subfigure[Gradient$\times$input attributions for digits 3 and 6]{
    \includegraphics[width=\columnwidth]{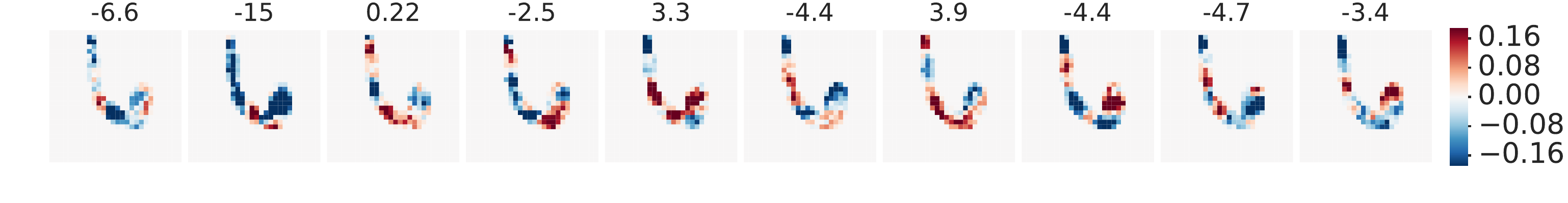}\label{fig:explanation_b}
    }
    \caption{Example attribution maps. 
    We train a linear Mnist classifier with bias parameters set to zero, $f_c(\x)=\sum_k w_{ck}x_k$. (An extra soft-max operation and cross-entropy loss were used for training.)
    (a) Gradients $\dxy{x_k}{f_c}$ are shown for each of the classifier outputs $f_c$ (coding for digits 0--9) and every pixel $k$. In a linear system, gradients are constant ($\dxy{x_k}{ f_c}=w_{ck}$).
    (b) For a given $\x$, ``gradient$\times$input''  attributions ($\ax_{ck}\equiv w_{ck}x_k$) decompose the output into components due to every input dimension, here shown for example digits  3 (middle) and 6 (bottom row). This decomposition is complete:  $f_c=\sum_k \ax_{ck}$, $f_c$ is shown over each plot. 
    \label{fig:explanation}
     }
\end{figure}
The task we study is image classification.
Importantly, we focus on explanations specific to a given network. Thus, our work is relevant for deciding which networks are safe to deploy, rather than studying image classification itself (cf \citet{RiegerKai2019, KindermansEtAl2017}).

We distinguish two types of visual explanations: saliency maps and attribution maps. For a given RGB image, a \emph{saliency map} highlights which regions are important for the classification; its values are non-negative. An \emph{attribution map} emphasises the sign of the evidence,  i.e.\ whether a given region adds value to the output, or whether its contribution is negative \cite{Ancona2017}. For any attribution map, its absolute value can be used as a saliency map.
Both saliency and attribution maps match the size of the image, losing the colour channel detail. This simplifies the visualisation, allowing
to overlay the images and their explanations, fostering interpretability. Attribution maps are often visualised with \emph{heatmaps}, using the topology of the input space and colour coding, as in Fig.\,\ref{fig:explanation}. 

Two main approaches emerged in extracting visual explanations. Some scientists focus on the image space, using occlusion and masking as the means of estimating saliency maps \cite{Zhou2014, ZeilerFergus2014, Fong2017, Dabkowski2017}. 
Others focus on the model output for a given class, devise rules for decomposing it and back-propagating the decomposition to the input space \cite{BaehrensEtAl2010, SimonyanEtAl2014, BachEtAl2015,  SelvarajuEtAl2016, SundararajanEtAl2017, SmilkovEtAl2017, Shrikumar2017, Srinivas2019}. 
%
%
In fact, the first attribution and saliency maps were based on the simplest gradient back-propagation \cite{BaehrensEtAl2010, SimonyanEtAl2014}. Input gradients, however, are not sufficient to recover the class output, a task explored by a range of attribution methods (see \citet{SundararajanEtAl2017, lundberg2017} for reviews). 

In this study, we investigate one of the simplest attribution methods: ``gradient$\times$input''. In many neural networks, e.g. made of rectified linear units (ReLU), ``gradient$\times$input'' attributions refer to the exact elements of computation. However, in order to recover the class score exactly, 
one needs to add other components, such as bias attributions. Such a ``full gradient'' decomposition of a class score has been recently described by \citet{Srinivas2019}. As they point out, input attributions can provide the exact decomposition only in a system with no bias parameters (Fig.\,\ref{fig:explanation_b}). 

Motivated by the ``full gradient'' decomposition, \citet{Srinivas2019} propose a FullGrad method of extracting visual explanations. The authors claim that adding attributions derived from bias parameters benefits the visual explanations.
%
However, most saliency methods so far have ignored the role of biases. 
In this work, we ask: How justified is it to ignore bias parameters in image classification? Is the exact decomposition of class score important for visual explanations? 

In order to address these questions, we develop a new procedure for evaluating pixel-based explanations. Our method emphasises correct classification and discriminatory power of visual explanations. We apply statistical tests when comparing different saliency methods. 

The structure of the paper is as follows:
In \S\ref{sec:background}, we introduce the notation, describe the FullGrad method and a way to modify it. In \S\ref{sec:impact_of_biases}, we evaluate the impact of bias parameters in over 20 ImageNet classifiers. In two of these networks, we decay bias parameters in order to test whether the exact decomposition of class score improves visual explanations. 
In \S\ref{sec:evaluating_expectations}, we develop a procedure for evaluating pixel-based explanations.
In \S\ref{sec:choosing_best_explanations}, we evaluate explanations derived from hidden activities and compare them with explanations derived from biases. While FullGrad relies on bias attributions, we propose to use attributions from hidden activities instead. 
We show our approach produces superior saliency maps.

\section{Background}\label{sec:background}

\subsection{Output decomposition via attributions}\label{subsec:basic_notation}
In a standard classification setting, an input $\x$ is fed to the classifier, which produces class scores $f_c(\x)$ for all possible classes $c\in(1,\ldots C)$. The maximum score points to the winning class. 
For a linear classifier, class scores are computed as a linear mapping over $\x \in \mathcal{R}^K$:
\begin{equation}
f_c(\x)=\sum_{k=1}^K w_{ck}x_k,\label{eq:linear_system}
\end{equation}
In such a classifier, input gradients recover the weights: 
\begin{equation}
\dxy{x_k}{f_{c}(\x)}=w_{ck}.
\end{equation}
Gradient visualisations (Fig.\,\ref{fig:explanation_a}) provide insight into how the linear system works, by quantifying how each class weights every pixel. 
However, in order to provide a \emph{complete} decomposition of the class output for a given $\x$, i.e.\ one which satisfies:
\begin{equation}
    f_c(\x) =\sum_{k=1}^K \ax_{ck}(\x)\label{eq:completeness}
\end{equation} 
we define $\ax_{ck}$ as \emph{gradient$\times$input} attributions:
\begin{equation}
\ax_{ck}\equiv x_k\dxy{x_k}{f_c(\x)}.\label{eq:attribution_x}
\end{equation} 

Equation (\ref{eq:completeness}) is a starting point of many attribution methods\footnote{Completeness is sometimes defined in reference to an arbitrary baseline $x_0$:  $f_c(\x)-f_c(\x_0) =\sum_{k=1}^K \ax_{ck}(\x, \x_0)$  \cite{Shrikumar2017, SundararajanEtAl2017}. Our formulation is stronger, as  setting $\ax_{ck}(\x, \x_0)\gets \ax_{ck}(x)-\ax_{ck}(x_0)$ meets that criterion.}. In a linear system, attributions $\ax_{ck}$ (Eq.\,(\ref{eq:attribution_x})) relate exactly to the computation performed by the classifier along each $k$.
For a grey image, gradient$\times$input attributions can be treated directly as attribution maps  (Fig.\,\ref{fig:explanation_b}). For coloured images, it is common to sum the components over colour channels:
\begin{equation}
\ax_{cij}\equiv \sum_{\phi\in(\mathrm{R,G,B})} x_{ij\phi}\dxy{x_{ij\phi}}{f_c(\x)}\label{eq:attribution_x_rgb},
\end{equation} 
with tuple $(i,j,\phi)$ coding for spatial dimensions and colour.

Attribution maps $\ax_{cij}$ inform us about the concrete evidence for a given image $\x$. In examples shown in Fig.\,\ref{fig:explanation_b}, pixels marked in red contribute positively to the classifier output. For example, nearly all pixels corresponding to digit 3 are marked red for the output of that class ($f_3$, fourth column). In contrast, most pixels in the image of 6 contribute negative evidence to $f_1$ (second column).

\subsection{Output decomposition in ReLU networks}\label{subsec:relu}
\begin{figure*}[t]
    \centering
    \includegraphics[width=2\columnwidth]{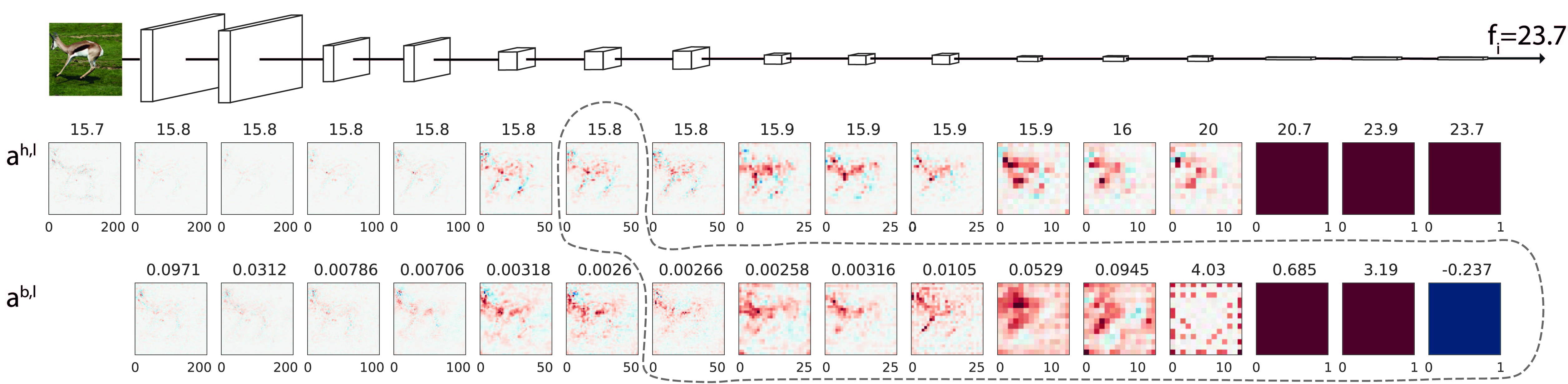}
    \caption{Complete decomposition of a ``gazelle'' class output in \vgg. An image of a gazelle is fed into the \vgg\ classifier, which outputs $f_c=23.7$ in favour of that class. Per layer activity attributions ($\ahl$) and bias attributions ($\abl$) are shown in color (red positive, blue negative), with their gross sum ($A^{h,l}_c$, $A^{b,l}_c$) displayed over each panel.
    The class score can be decomposed in many ways, the dashed line highlights one of them (Eq.(\ref{eq:alternative_decompositions_A})):
    $f_c=A^{h,l=6}_c + \sum_{l^\prime=7}^{16}A^{b, l^\prime}_c$.
    \label{fig:attribution_vgg_example}}
\end{figure*}
The intuition derived from the linear system applies to a range of neural networks. The most common ReLU networks afford piece-wise linear mappings. 
Importantly, although the first visual explanations were developed for probabilistic models \cite{BaehrensEtAl2010}, the authors recognised the problem of gradients diminishing for high-certainty decisions. To circumvent the problem, \citet{SimonyanEtAl2014} suggested to analyse score (i.e.\ ``logits'') rather than probabilities, which is the approach we take here. 

For a ReLU network with no bias parameters, which we refer to as a \emph{zero-bias} network, we can write as in Eq.\,(\ref{eq:linear_system}):
\begin{equation}
f_c(\x)=\sum_{k=1}^K w_{ck}(\x)x_k\label{eq:zero_bias_network}
\end{equation}
Thus, in a zero-bias network, a complete decomposition of class score is possible using input attributions alone (Eqs (\ref{eq:completeness}) and (\ref{eq:attribution_x}) combine to Eq.\,(\ref{eq:zero_bias_network})). 
Note that the weights depend on the input, which means they need to be evaluated separately for every $\x$, e.g. using gradient back-propagation.

In general, a complete decomposition of a class score requires taking bias parameters into account. Defining bias attributions as $a^b_{cn}\equiv b_n\dxy{b_n}{f_c(\x)}$, \citet{Srinivas2019} write out the ``full gradient'' decomposition: 
\begin{equation}
f_c(\x) =\sum_{k=1}^K \ax_{ck} + \sum_{n=1}^N a^b_{cn}\label{eq:full_gradient}.
\end{equation}   
This decomposition is a starting point of the FullGrad method. For each convolutional layer $l$, and feature $\phi\in\Phi^l$, FullGrad  combines the different attributions into a single saliency map. Replacing index $n$ with $(l, \phi, i, j)$ and omitting the spatial coordinates for clarity, they get:
\begin{equation}
\s \gets \Psi(\mathbf{\ax}_{c}) + \sum_{l=1}^L\sum_{\phi\in\Phi^l} \Psi(\mathbf{a}^{b,l}_{c\phi}),\label{eq:FullGrad}
\end{equation}
The transformation $\Psi(\cdot)$ needs to match the sizes of the different attributions to that of the input image. \citet{Srinivas2019} recommend using:
\begin{equation}
\Psi(\a)=\mathrm{bilinearUpscale}(\mathrm{rescale}(\abs(\a)))
\end{equation}
where each feature map is normalised with:
\begin{equation}
\mathrm{rescale(\a)} = \frac{\a-\min(\a)}{\max(\a)-\min(\a)}.\label{eq:rescale}
\end{equation}
According to \citet{Srinivas2019}, FullGrad with their choice of $\Psi(\cdot)$ produces superior saliency maps.

\subsection{Alternative decompositions}\label{subsec:alternative_decompositions}
``Full gradient'' (Eq.\,(\ref{eq:full_gradient})) is a well justified and complete decomposition of the class score. However, similar decompositions can be obtained for any layer in the network. 
Others have shown the benefit of extracting saliency maps from hidden activities \cite{Zhou2015, SelvarajuEtAl2016}. Below, we extend the ``full gradient'' formalism to the entire depth of the network.

Let us generalise the notion of input attributions (Eq.\,(\ref{eq:attribution_x_rgb})) to any network layer $l\in(1,L)$. We define hidden activity attributions $\ahl$ and bias attributions $\abl$ per spatial position ($i,j$) of neurons in layer $l$, and the class $c$ considered:
\begin{eqnarray} 
a^{h,l}_{cij}&\equiv \sum_{\phi\in\Phi^l} h^{l}_{ij\phi}\dxy{h^l_{ij\phi}}{f_c}\label{eq:attribution_hl}\\
a^{b,l}_{cij}&\equiv \sum_{\phi\in\Phi^l} b^{l}_{ij\phi}\dxy{b^l_{ij\phi}}{f_c}\label{eq:attribution_bl}
\end{eqnarray} 
with  $a^{h,0}_{cij}\equiv \ax_{cij}$ (Eq.(\ref{eq:attribution_x_rgb})).
We can now write $L+1$ complete decompositions, for $l\in(0,\ldots L)$:
\begin{eqnarray}
f_c&=&\sum_{ij} a^{h,l}_{cij} + \sum_{l^\prime=l+1}^{L} \sum_{ij} a^{b,l^\prime}_{cij}\label{eq:alternative_decompositions}.
\end{eqnarray}  
Note, that for $l=0$, we recover the original ``full gradient'' formulation,  Eq.\,(\ref{eq:full_gradient}). (For exact equations in case of ReLU networks, see Supplementary \S\ref{sec:app:relu_decomposition}.)

We illustrate $\ahl$ and $\abl$ in \vgg, for an example image of a gazelle (Fig.\,\ref{fig:attribution_vgg_example}). From left to right, attributions decrease in size, reflecting the size of the corresponding convolutional layers (sketched on top). The number over each panel is a gross sum of attributions per layer: \begin{eqnarray}
A^{h,l}_c\equiv\sum_{ij} a^{h,l}_{cij}, &
A^{b,l}_c\equiv\sum_{ij} a^{b,l}_{cij}\label{eq:gross_sum}
\end{eqnarray}
The dashed line highlights which attributions need to be summed to obtain an example decomposition of the class score (Eq.\,\ref{eq:alternative_decompositions}), in this case for class $c=\mathrm{gazelle}$:
\begin{equation}
f_c = A^{h, l}_c + \sum_{l^\prime=l+1}^L A^{b, l^\prime}_c.\label{eq:alternative_decompositions_A}
\end{equation}
Most $\ahl$ and $\abl$ seem to focus over the gazelle in the picture, except for the last convolutional layer, where $\a^{b}_{l=13}$  is clearly artefactual and strongest around the edges.

\subsection{Our approach and related work}\label{subsec:our_method}
Inspired by the superior quality of $\ahl$ over $\abl$ attributions (Fig.\,\ref{fig:attribution_vgg_example}), we propose to use them in place of $\abl$ in an approach similar to FullGrad: 
\begin{equation}
\s\gets \Psi(\mathbf{\ax}) + \sum_{l=l_0}^L \Psi(\ahl).\label{eq:OurApproach}
\end{equation}
We choose $l_0$ empirically, deploying our evaluation method (\S\ref{sec:evaluating_expectations}) to decide how many layers to aggregate over.

Note that we apply $\Psi(\cdot)$ to $\ahl$, rather than to every feature map (cf Eq.\,(\ref{eq:FullGrad})). Thus, our approach emphasises attributions per layer, rather than per feature. Commonly, the number of features grows for spatially smaller layers (Fig.\,\ref{fig:attribution_vgg_example}\ top), which means FullGrad effectively over-emphasises low-resolution visual explanations. 

We should note the difference between our approach and gradCAM \cite{SelvarajuEtAl2016}, the only other method which extracts explanations from deep layers; Rather than $\ahl$ (Eq.\,(\ref{eq:attribution_hl})), gradCAM computes $\sum_\phi h^{l}_{ij\phi} \sum_{ij} \dxy{h^l_{ij\phi}}{f_c}$, thus summing the gradients separately for each feature $\phi$. The authors use rectification, rather than an absolute value to generate saliency maps and conclude that the top layer yields the best looking visualisation. Instead of aggregating saliency maps over the layers, they use Guided Backprop  \cite{Springenberg2014} to sharpen their visualisation.

In fact, it is common for saliency methods to devise arbitrary rules \cite{Springenberg2014, ZeilerFergus2014, BachEtAl2015}. Although unifying frameworks have been proposed \cite{Ancona2017, lundberg2017}, a normative theory of what makes a good visual explanation is still missing.
The ``full gradient'' approach stands out as a principled justification for the choice of attributions: considering the exact components of the computation. Another approach motivated by completeness and uniqueness, Integrated Gradients \cite{SundararajanEtAl2017}, also yields gradient$\times$input attributions, albeit in an differential form. In addition, Integrated Gradients rely on the choice of a baseline $\x_0$, which strongly affects the saliency maps (see \citet{Srinivas2019} for a discussion). 
 
\section{The impact of biases in image classification}\label{sec:impact_of_biases}
While our choice of attributions (``gradient$\times$activity'') aligns with the ``full gradient'' perspective, it deviates from the assumption of completeness (Eq.\,\ref{eq:completeness}).
Thus, it is most relevant to ask: Does completeness matter? Do biases matter? Will eliminating them improve the visual explanations?
\begin{figure}
    \includegraphics[width=\columnwidth]{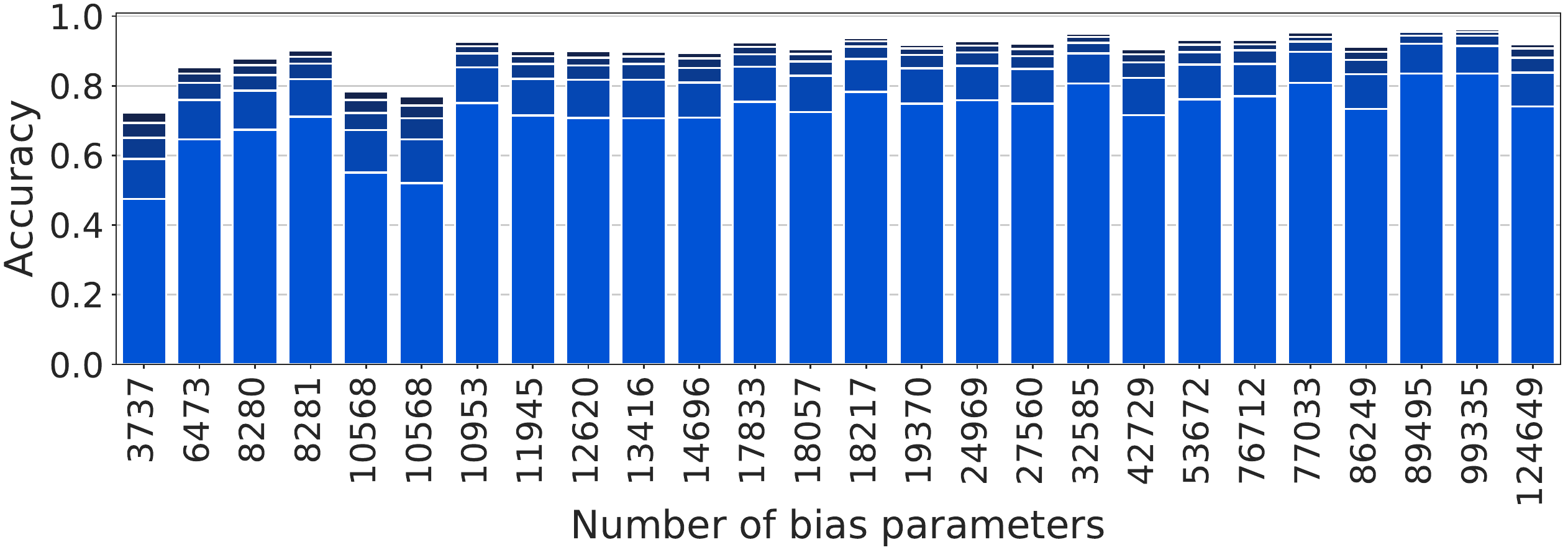}
    %
    \includegraphics[width=\columnwidth]{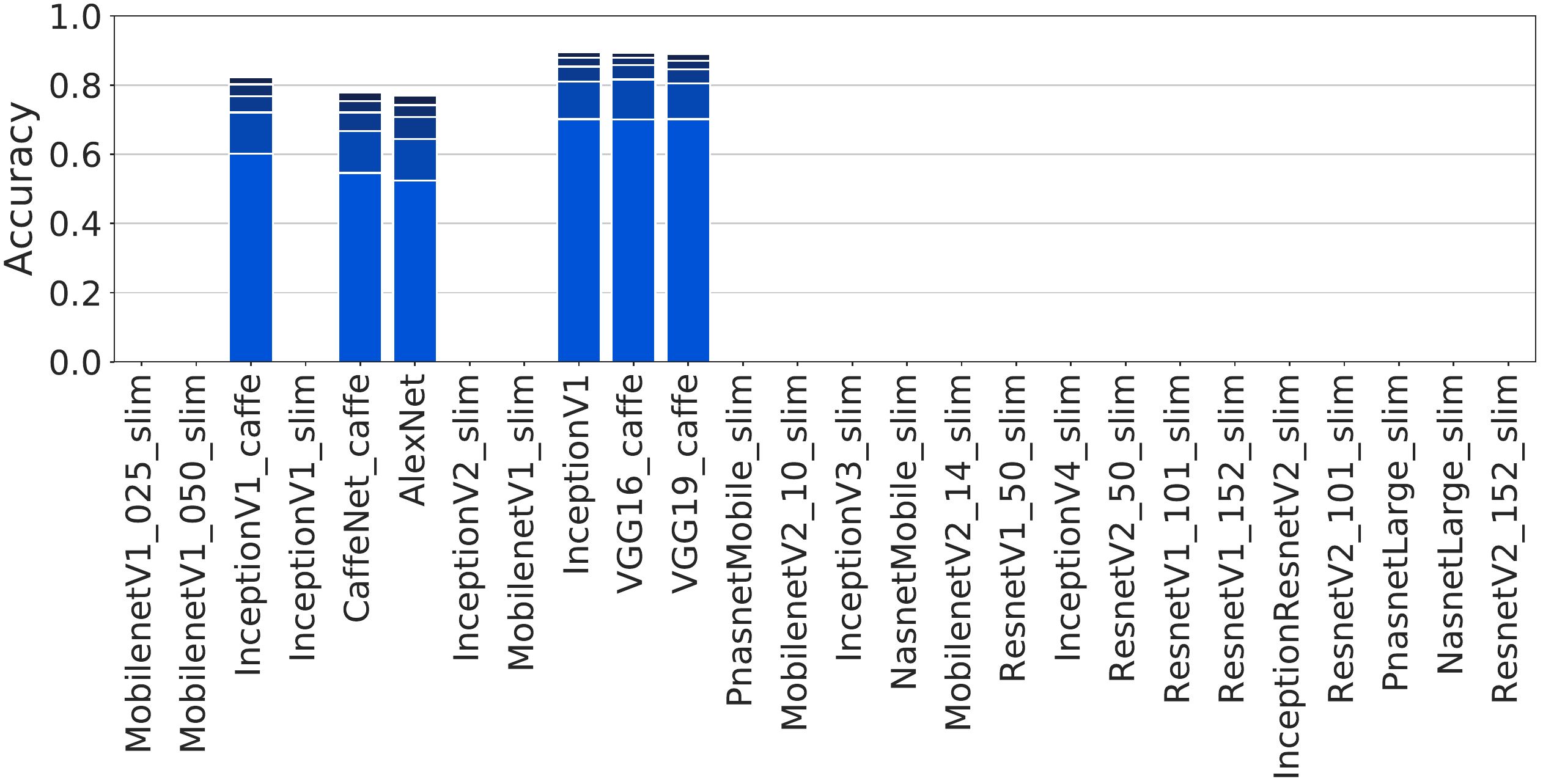}
    \caption{Networks available in the Lucid package.
    Test accuracy of networks available in the Lucid package before (top) and after setting bias parameters to zero (bottom). The models are arranged according to the number of bias parameters. The total length of the bar denotes top-5 accuracy, while the white segments indicate top-1 to top-4 accuracy. 
    All  networks sensitive to bias removal use Batch Normalization.
    Larger networks achieve higher accuracy.
    \label{fig:lucid}}
\end{figure}

\begin{figure*}
    \centering
    \subfigure[Vanilla \resnet]
    {
    \includegraphics[width=2\columnwidth]{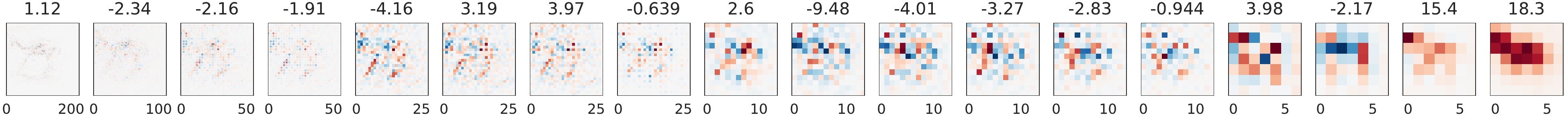}\label{fig:attribution_resnet_example_a}
    }
    \subfigure[\resnet\ with biases set to zero]
    {
    \includegraphics[width=2\columnwidth]{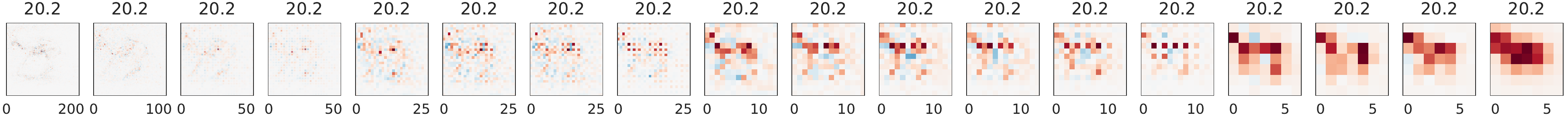}
    }
    \caption{Activity attributions in \resnet.  $\ahl$ are shown for the same gazelle image as in Fig.\,\ref{fig:attribution_vgg_example}, for $l=0$ (input attributions, left) to $l=17$ (right). Their gross sums $A^{h,l}_c$ shown over each heatmap point at the undisguised role of biases ($\abl$). In the original \resnet, $A^{h,l}_c$ are negative for many layers. After removing biases and retraining the network, $\ahl$ provide a complete decomposition of the network output (20.2) and correlate better with the position of the gazelle across all layers.}
    \label{fig:attribution_resnet_example}
\end{figure*}

We surveyed networks trained on ImageNet and available via the Lucid package \cite{lucid}
to estimate how bias parameters contribute to net performance. In Figure \ref{fig:lucid}, the networks are arranged by the number of bias parameters. Best performing networks do not require the highest number of biases (e.g. InceptionV2\slim).

Removing biases destroys the performance of most of the modern image classifiers (Fig.\,\ref{fig:lucid}, bottom), except for AlexNet, CaffeNet, VGG16/VGG19, InceptionV1 (GoogleNet) and Inception V1\caffe, which are the only networks that do not use Batch Normalization (\bn). In fact, without biases, only CaffeNet suffers a drop in performance larger than 1\% (\SFig.\,\ref{fig:app:accuracy_zero_bias_diff}). 

For a more detailed analysis of the role of the bias, we choose two ReLU networks: \vgg\ \cite{SimonyanZisserman2014} and \resnet\ \cite{HeEtAl2016}. %
We decay bias parameters to obtain zero-bias versions of these models (Supplementary \S\ref{sec:app:bias_decay}). Upon fine-tuning to match the outputs of the original network, we see a full recovery of performance in \vgg\ (Supplementary Fig\,\ref{fig:app:decay_vgg}) and  $>$99\% recovery in the \bn-trained \resnet\ (Supplementary Fig\,\ref{fig:app:decay_resnet}).

In \vgg, setting biases to zero has little effect on visual explanations (Supplementary Fig.\,\ref{fig:app:correlations_vgg}). We thus explore the impact of completeness on visual explanations in \resnet\ (Supplementary Fig.\,\ref{fig:app:correlations_resnet}). 
Activity attributions derived before and after decaying bias parameters are presented in  Figure \ref{fig:attribution_resnet_example}. 
As in Fig.\,\ref{fig:attribution_vgg_example}, the numbers over each $\ahl$ heatmap represent the per-layer sum, $A^{h,l}_c$ (Eq.\,\ref{eq:gross_sum}). In the vanilla \resnet, the $A^{h,l}_c$ values differ greatly from the final network output (18.3), indicating the role of missing bias attributions (not shown here). In contrast, $A^{h,l}_c$ in the zero-bias network are all identical and equal to the class score in that network (20.2), as expected from a complete decomposition (Eq.\,\ref{eq:zero_bias_network}).
Does the quality of visual explanations improve with completeness?

A closer look at the $\ahl$ attributions reveals they tend to swap sign, i.e.\ large image regions turn blue (negative, Fig.\,\ref{fig:attribution_resnet_example_a}). We hypothesise the reason for this difference: Rectification restricts the choice of the sign for the weights in a zero-bias system. For example, the simplest identity function can be represented in a ReLU network by $y=w^{-1}[wx + b]_+ + b/w$ for both positive and negative $w$ (as long as $b$ is suitably large to assure $wx + b > 0$ across the entire input range). Without biases, $y=w^{-1}[wx]_+$ is non-zero only for a positive $w$.

In Supplementary \S\ref{sec:app:saliency_biasfree}, we provide a quantitative comparison of the visual explanations extracted from vanilla and zero-bias networks. Indeed, attribution maps from  zero-bias \resnet\ are more accurate (Supplementary Fig.\,\ref{fig:app:perturbation_bias_free_attributions}). However, the saliency maps do not improve much. Overall, decaying bias parameters did not enhance the best visual explanations that can be extracted from \resnet\ (Supplementary Figs \ref{fig:app:perturbation_averaging_over_layers} and \ref{fig:app:perturbation_bias_free}). Thus, we conclude that  completeness is not essential for interpretability. 

\section{Evaluating explanations: the method}\label{sec:evaluating_expectations}
We now explain our procedure of evaluating pixel-based explanations. It  enables us to choose between different methods of extracting visual explanations. 

Previous studies have primarily relied on image perturbation as a means for evaluating saliency maps. In the most common \emph{pixel removal} approach, image regions are masked while the class score is monitored \cite{SamekEtAl2016}. The masks are chosen according to the given saliency map $\s$. When starting from the most important pixels ($+\s$), one expects a greater change in the output. Starting from the least important ($-\s$), a smaller change is expected. One can also measure the gap between the two effects \cite{SamekEtAl2016}.

Unfortunately, perturbations have a major drawback: the modified images are likely to confuse the network in unexpected ways \cite{SamekEtAl2016, Fong2017, Dabkowski2017}. This is particularly evident in the pixel removal approach, which generates patterns with high spatial frequencies. Such patterns tend to destroy the network performance regardless of where they are positioned. It becomes difficult to tell a rational explanation from an obscure one, such as a random pixel removal. Several methods of perturbing pixels were explored by \citet{SamekEtAl2016}, but to our best knowledge no method exists that addresses this issue directly. 
The method we propose yields results consistent for both $+\s$ and $-\s$ perturbation approaches, and comes with highly significant p-values.

We propose to modify the evaluation procedure: Instead of recording how much the class score $f_c$ changes with the number of perturbed pixels, we increase the perturbation until the modified image $\x_s$ changes the classification, i.e.\ $\mathrm{argmax}(f(\x_s))\neq\mathrm{argmax}(f(\x))$. 
We record the relative amount of signal which has to be removed from the image to change the decision:
\begin{equation}
\e(\x, \s) = 1-||\x_s||_2/||\x||_2,
\end{equation}
where $||.||_2$ stands for the Euclidean norm. 

For example, applying perturbations over pixels deemed as least important ($-\s$ approach), we expect to remove more signal without affecting the network decision. Thus, higher values of $\e(\x, -\s)$, should indicate a higher accuracy of our saliency map $\s$. 

In order to dissociate the meaningful explanations from accidental ones, we run multiple perturbations for a given image $\x$. We compare $\e(\x, \s)$ evoked by the valid saliency map ($\s$), to $\e(\x,\bxi)$ obtained from a random map ($\bxi$). Most importantly, we use patterns $\bxi$ which are drawn from the same marginal distribution as $\s$. To do this, we choose $\bxi$ from a pool of saliency maps extracted with the same method, but for images  $\x^\prime \neq \x$. This ensures that, at a distributional level, the $\bxi$ share the same statistics as the $\s$, e.g.\ having similar spatial frequency and pixel values. The relative effect of the two types of perturbations,
\begin{equation}
\delta \e = \e(\x, \s)-\e(\x, \bxi)
\end{equation} is a more accurate measure of how $\s$ is related to image classification. It estimates the discriminatory power of a given explanation relative to the pool of alternative explanations. In Supplementary \S\ref{sec:app:perturbations}, we show it yields consistent results, independent of the average effect of perturbations.

When comparing saliency methods, we advise to use pairwise tests, i.e.\  to compare the method effectiveness per image. This is because the perturbation metrics vary widely depending on the size of the region of interest.

In the following, we report our results using the metrics:
\begin{eqnarray}
\ee&\equiv&\e(\x, -\s)\\
\delta \ee &\equiv& \ee-\e(\x, -\bxi)\label{eq:delta_e}
\end{eqnarray}
which increase with the quality of the saliency map. Both will be of value when choosing best method to extract explanations. This is because we want our explanations to be not only the most discriminative (maximising $\delta \ee$), but also most comprehensive, i.e.\ least likely to perturb the network in unexpected ways (maximising $\ee$).

For a detailed evaluation of the perturbation methods and measures we tried, we refer the reader to Supplementary \S\ref{sec:app:perturbations}. 

\section{Evaluating explanations: results}\label{sec:choosing_best_explanations}
In Figure \ref{fig:attribution_vgg_example}, we made the qualitative observation that hidden activity attributions are superior to bias attributions for an example. We proceed to quantify this rigorously.
\begin{figure}
    \centering
    \includegraphics[width=.95\columnwidth]{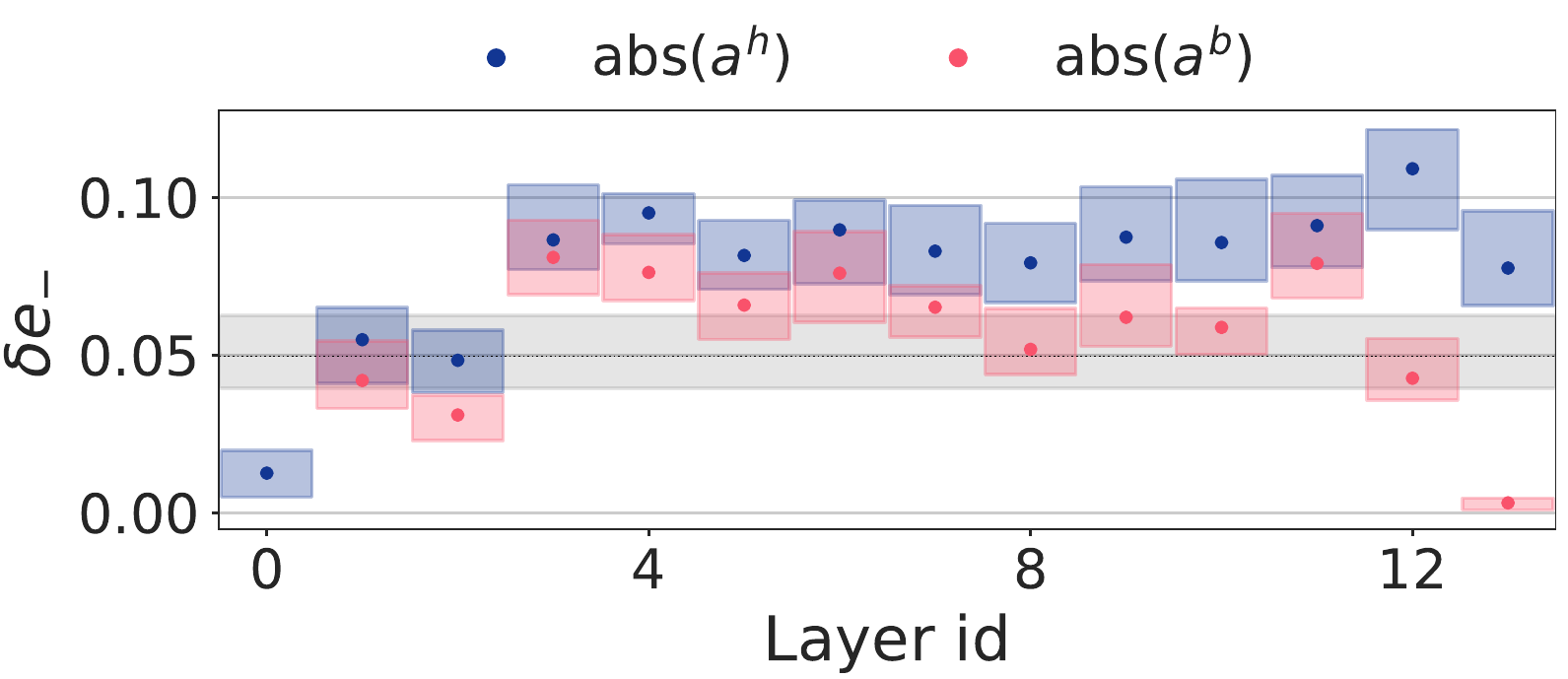}
    \caption{Activity attributions are more salient than bias-related attributions. In \vgg, we used layer-specific $\abs(\ahl)$ and $\abs(\abl)$ saliency maps to perturb the images (after bilinear up-sampling to match the input image size). Across all layers, hidden activity attributions more accurately than biases delineate the parts of the image important for the network. Input attributions ($0^{\text{th}}$-layer) are inferior to the standard gradients approach ($\s\gets\sum_{\phi\in(R,G,B)}\abs(\dxy{x_{k,\phi}}{f_c})$, shown in grey). Boxes represent 25-th to 75-th quantiles.}
    \label{fig:perturbation_vgg_ab}
\end{figure}

Using  $\Psi(\a)\equiv\mathrm{bilinearUpsample}(\abs(\a))$ to transform $\ahl$ and $\abl$ into saliency maps, we run perturbation experiments for each layer and attribution type. As shown in Fig.\,\ref{fig:perturbation_vgg_ab}, 
$\ahl$ (blue) provide more accurate saliency maps than $\abl$ (red) in every layer; p-values range between $p<10^{-5}$ ($l=$3) and $p<10^{-130}$ (last layer; Wilcoxon signed rank test).

\begin{figure}
    \centering
    \includegraphics[width=.99\columnwidth]{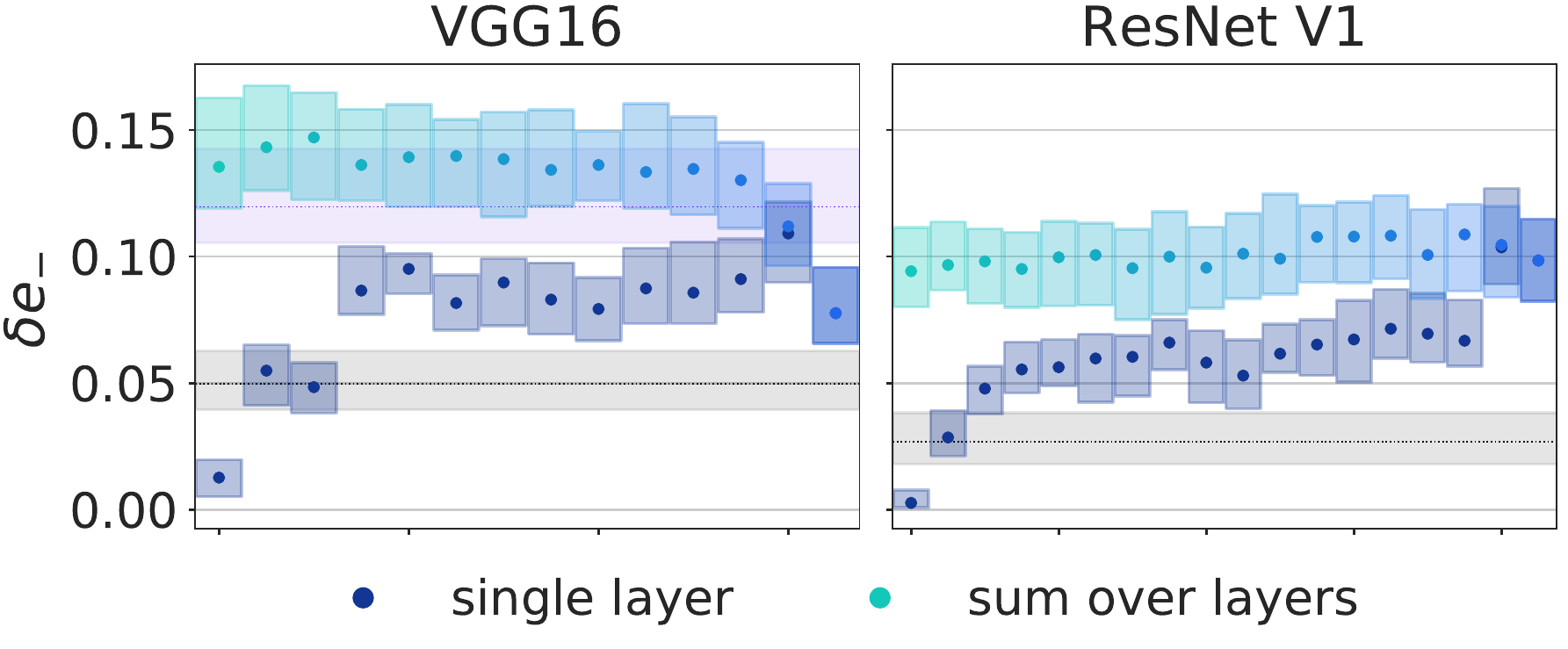}\\
    \includegraphics[width=.99\columnwidth]{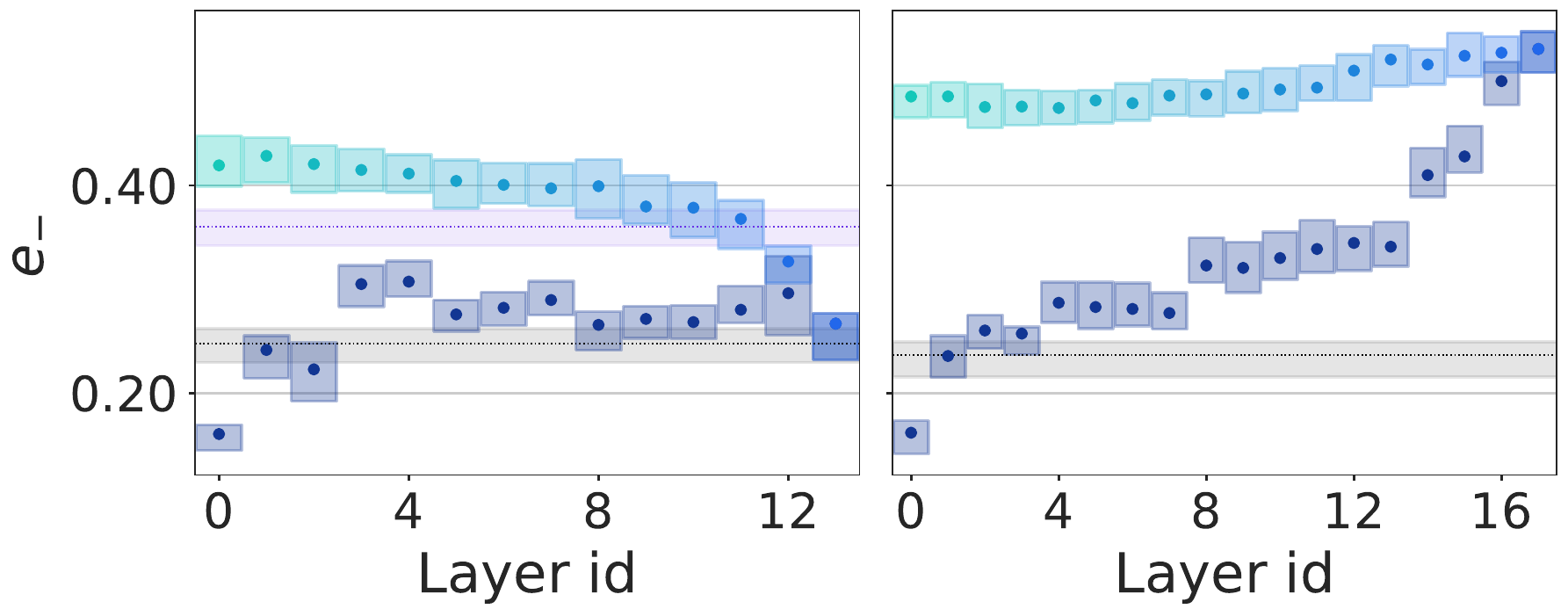}
    \caption{Aggregating over layers improves saliency in \vgg, less so in \resnet. We present both the $\delta \ee$ metric which increases for more discriminative saliency maps (top); as well as $\ee$, which is higher for better estimates of unimportant pixels (bottom). 
    Single layer explanations are shown in blue. Aggregate explanations are formed by averaging single-layer saliency maps from top to bottom (blue to cyan). For \vgg, we show ``FullGrad'' approach in addition to the input gradient baseline (purple line).  (Our implementation of ``FullGrad'' differs in that we aggregate bias attributions per layer, rather than per feature, see \S\,\ref{subsec:our_method}.) \\
    In \vgg, it is beneficial to average attributions over all layers, leaving out input attributions. In \resnet, the most accurate (top) and comprehensive (bottom) explanations are extracted by the average of the top two layers. }
    \label{fig:perturbation_averaging_over_layers}
\end{figure}
Can we capitalise on that difference?
As described in \S\,\ref{subsec:our_method}, we aggregate saliency maps extracted from $\ahl$ (Eq.\,(\ref{eq:OurApproach})), after squashing their values between 0 and 1 (Eq.\,(\ref{eq:rescale})).

Our results are presented in Fig.\,\ref{fig:perturbation_averaging_over_layers}. In addition to the metric presented before ($\delta\ee$, top), we also display the overall impact of the perturbation ($\ee$, bottom). As discussed in \S\ref{sec:evaluating_expectations}, we expect from a good explanation to be both discriminative (higher $\delta\ee$) as well as comprehensive (higher $\ee$).

The quality of single layers is shown in blue, cyan corresponds to a sum over saliency maps derived from the input and all the layers. The remaining colours represent top-down aggregates (from the given layer $l_0$, Eq.\,(\ref{eq:OurApproach})). For \vgg, we add the ``FullGrad'' baseline (purple line).

In \vgg, the sum of three top layers already supersedes the quality of the ``FullGrad'' saliency map. Taking more layers into account further improves the explanation, both its discriminative (Fig.\,\ref{fig:perturbation_averaging_over_layers} top, $p<10^{-5}$) and informative (bottom, $p<10^{-10}$) qualities. Interestingly, adding input attributions, the only component of the ``full gradient'' decomposition (Eq.\,\ref{eq:full_gradient}), hurts the visual explanations according to both $\delta\ee$ and $\ee$.

In \resnet, the penultimate layer appears to contain the most accurate information. However, summing the top two layers is similarly discriminative, but more comprehensive $\ee$ (bottom, $p<10^{-70}$). Adding more layers hurts both $\delta\ee$ and $\ee$, albeit the effect is not strong ($p<0.001$ for layers 9 and below). Thus, for \resnet, the best method of extracting explanations is to average the saliency maps $\abs(\ahl)$ from the top two layers of the network. 
Note, that this approach is better than focusing on the top layer, as advocated by CAM and gradCAM methods \cite{Zhou2015, SelvarajuEtAl2016}.

\begin{figure}
\subfigure[\vgg]{
\includegraphics[width=.99\columnwidth]{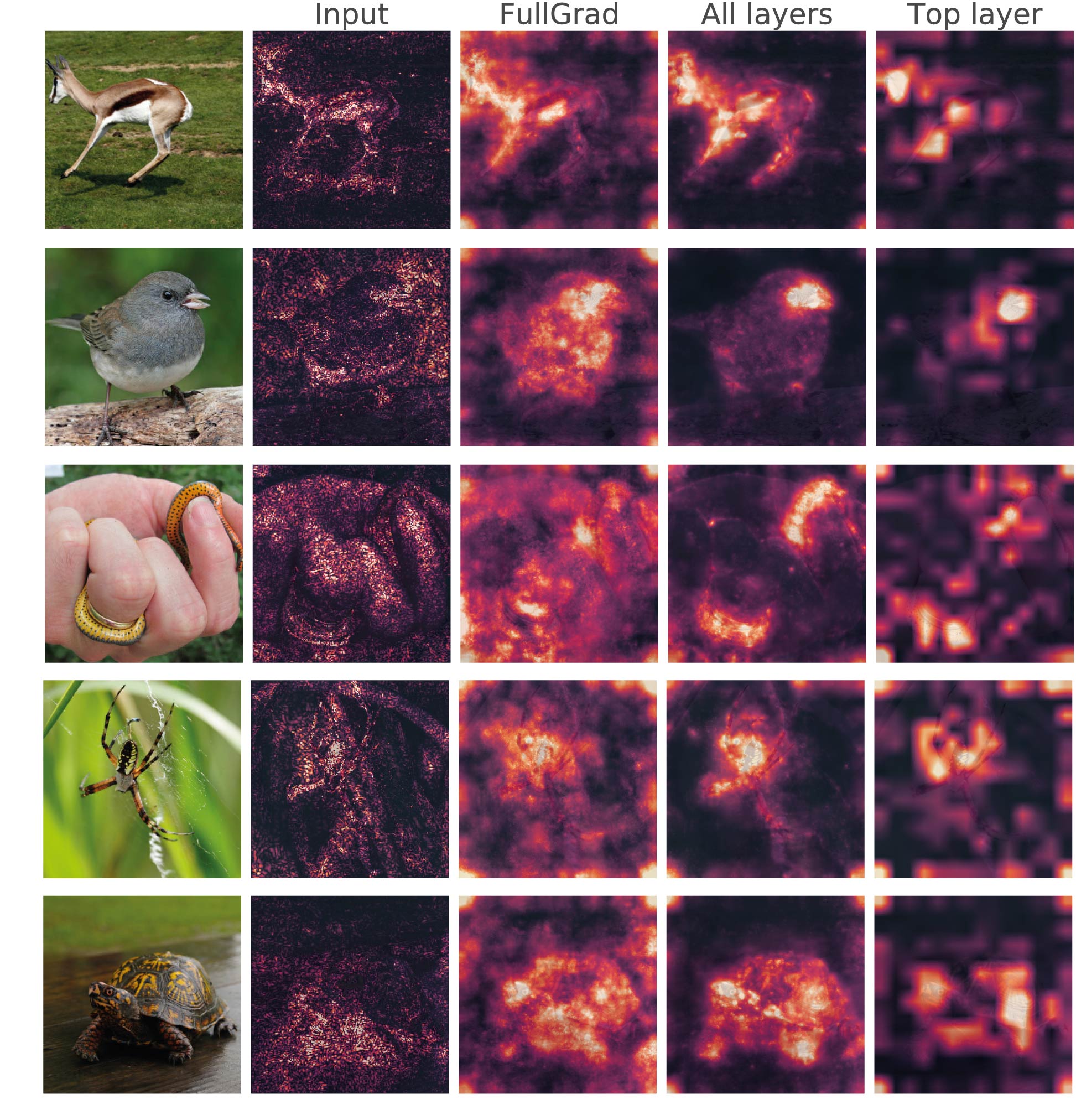}
}
\subfigure[\resnet]{
\includegraphics[width=.99\columnwidth]{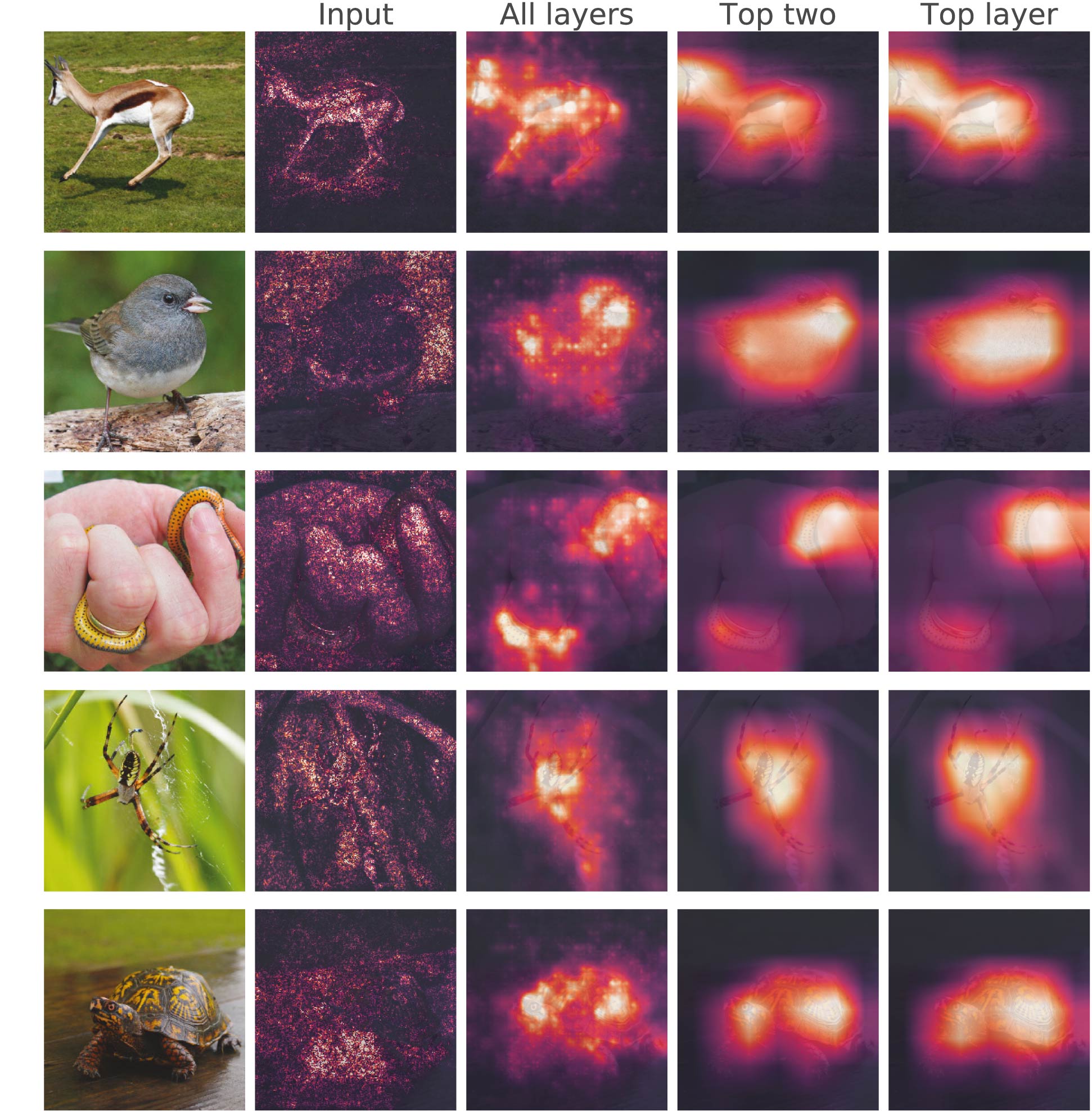}
}
\caption{Example saliency maps for methods chosen with our evaluation. The most informative saliency method for \vgg\ is ``All layers'':  averaging $\abs(\ahl)$ from all but the first (input) layer. For \resnet, averaging top two layers is most discriminative and comprehensive. Related approaches: input attributions, ``FullGrad'' and ``Top layer'' are shown for reference.}\label{fig:saliency_examples}
\end{figure}

All these quantitative observations can be confirmed by visual inspection (Fig.\,\ref{fig:saliency_examples}). All methods of explanation extraction are unquestionably superior to the input attributions. The difference to ``Top layer'' is more subtle in \resnet\ than in \vgg, but also discernible. In \vgg, the ``All layers'' approach yields cleaner explanations, more faithfully reflecting the interesting image content than either ``FullGrad'' or ``Top layer''. In \resnet, the ``Top two'' approach, albeit lower resolution than ``All layers'', seems more accurate in delineating important features. Thus, our procedure of evaluating the quality of explanations proves to be consistent with an intuitive judgement.

\section{Discussion}
Since the first study by \citet{BaehrensEtAl2010}, methods of extracting visual explanations ignored the role of bias parameters. Recently, \citet{KindermansEtAl2017} exposed a problem with this approach: explanations differ in models that have the same functionality, but different biases. One way to address this problem is to take biases into account when constructing explanations: an approach advocated by \citet{Srinivas2019}, the authors of the FullGrad method. Another solution is to set the biases to zero, thereby enabling a complete decomposition of network output by input attributions alone---an approach we tested here.

However, our results indicate that completeness has little impact on interpretability. Removing biases from \resnet\ prevents the single layer attributions from swapping sign, but does not improve the best explanations one can extract from this network. 
In fact, we discovered that biases do not play an important role in modern image classifiers. Although setting these parameters to zero incapacitates networks trained with \bn, our experiments suggest that their role is mostly superficial. With some extra training and a gentle decay procedure, we succeeded to keep $>99\%$ of \resnet\ performance in its zero-bias version.

Thus, we attribute the success of the FullGrad method to aggregating explanations across the network, rather than using an exact decomposition of network output.
Further, we show that aggregating hidden activity attributions, rather than bias attributions, produces superior saliency maps. This is good news especially for the modern image classifiers, which have a large number of biases ($\beta$ and $\mu$ parameters in \bn), making it difficult to correctly compute all the attributions. In contrast, hidden activity attributions are relatively easy to compute.

One may wonder why it is possible to extract explanations from hidden layers? Clearly, the local connectivity, which keeps a relatively consistent topology across layers, is paramount. We are optimistic our insight will apply not just to convolutional networks, but other systems with local connectivity.

Interestingly, in \vgg, it is beneficial to aggregate attributions from all network layers. In \resnet, the advantage is limited to the top layers. We hypothesise this difference is related to the choice of down-sampling. For the sake of interpretability, we recommend to use max-pooling  (\S\ref{subsec:app:perturbations_noise}). 
Further, we discovered that \bn\ decreases robustness to scaling and shifting input values in ImageNet classifiers, which likely affects their interpretability (\S\ref{subsec:app:scaling}).
A more detailed study is needed to estimate the impact of other factors, such as the presence of skip connections, convolutional layer sizes, and the use of drop-out.

Our detailed comparison of methods for extracting visual explanations was only possible thanks to our new evaluation protocol. Our procedure does not require special datasets \cite{Yang2019}, manipulation to the model \cite{AdebayoEtAl2018, Hooker2019}, or a human in the loop. 
\section*{Acknowledgments}
We thank Neil Rabinowitz and Joel Veness for valuable comments on the manuscript.
\bibliographystyle{icml2020}
\bibliography{saliency}

\appendix
\newpage
\setcounter{figure}{0}   
\setcounter{equation}{0}   
\twocolumn[
\icmltitle{Supplementary Material for ``Measuring and improving the quality of visual explanations''}
\vskip 0.3in
] 

\section{Example ReLU computation}\label{sec:app:relu_decomposition}
Defining attributions for hidden activities of a penultimate layer as $a^{h, L-1}_{ck}=w^{L}_{ck} [h^{L-1}_k]_{+}$, we write the class score as: 
\begin{eqnarray}
f_c &=& \sum_{k\in K^{L-1}} w^{L}_{ck} [h^{L-1}_k]_{+} + b^L_c.
\end{eqnarray}
Here, $w_{ck}^L$ and $b_c^L$ are the weights and biases of class $c$,  $[\cdot]_+$ denotes the rectification, and $h_k^{L-1}$ is the hidden activity of $k$-th neuron in the layer below:
\begin{equation}
h^{L-1}_k \equiv \sum_{k'\in K^{L-2}} w^{L-1}_{kk'} [h^{L-2}_{k'}]_{+} + b^{L-1}_k.
\end{equation} 

For the penultimate layer, the decomposition becomes:
\begin{eqnarray}
f_c &=&\sum_{k \in K^{L-1}_+, k'} w^L_{ck} w^{L-1}_{kk'} [h^{L-2}_{k'}]_+ \nonumber \\
&&+ \sum_{k \in K^{L-1}_+} w^L_{ck} b^{L-1}_k + b^L_c,
\end{eqnarray} 
where $K^{L-1}_{+}$ is the set of indices of $L-1$-layer neurons that were not rectified ($h^{L-1}_k>0$). 

Thus, the linear factors with which $h^{l}_k$ contribute to the output can be obtained by multiplying the ``active'' weights from layers above. For example, for $l=L-2$ above: 
\begin{equation}
\overline{w}^{L-2}_{ck'} \equiv \sum_{k \in K^{L-1}_+} w^L_{ck} w^{L-1}_{kk'}.
\end{equation}
For biases, a pattern starts to emerge, in which a complete decomposition at layer $l$ requires adding bias-related components from all layers above. For any layer, starting from the input, $l\in(0, L-1)$:
\begin{equation}
f_c= \overline{w}^{l}_{ck} [h^l_{k}]_+ + \sum_{l'=l+1}^L \overline{w}^{l'}_{ck'} b^{l'}_{k'}
\end{equation} 
with:
\begin{eqnarray}
\overline{w}^{l}_{ck^{l}} &\equiv& \sum_{l'=l+1}^{L-1}\sum_{k^{l'} \in K^{l'}_+} \left(w^L_{ck^{L-1}}\prod_{l'=l}^{L} w^{l'+1}_{k^{l'+1}k^{l'}}\right)\nonumber\\
\overline{w}^{L}_{ck} &\equiv& 1
\end{eqnarray} 
Note, that $\overline{w}^{l}_{ck^{l}}$ can be computed with a single call to back-propagated gradients:
\begin{equation}
\overline{w}^{l}_{ck^{l}} = \partial_{h^{l}_{k^l}} f_c
\end{equation}

\section{The role of bias parameters in ImageNet classifiers}\label{sec:app:the_role_of_bias}
In this section, we provide more details from our investigation of the impact of bias parameters in ImageNet classifiers.
\subsection{Robustness to scaling and shifting the input distribution}\label{subsec:app:scaling}

\begin{figure}
    \includegraphics[width=\columnwidth]{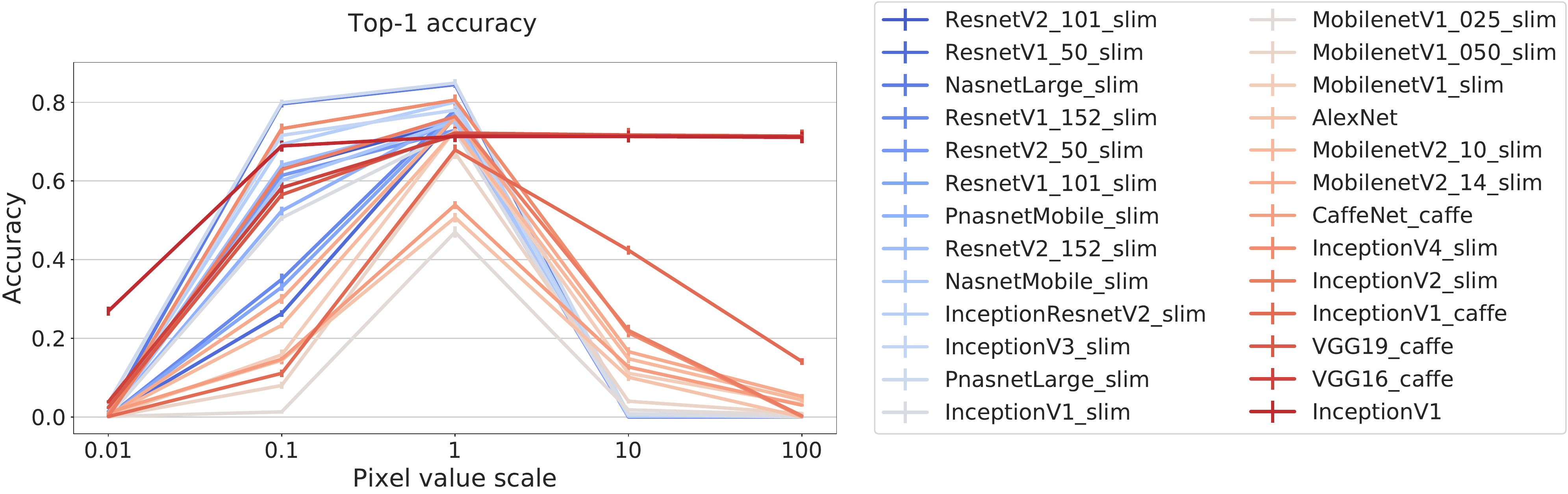}
    \caption{Robustness of modern image classifiers to changes in brightness. We modify ImageNet test set by scaling the range of pixel values by 0.001 up to 1000 times. We report top-1 performance, using colour according to the relative performance change at 1000\% of pixel values. 
    Only \vgg, VGG19 and InceptionV1 (``GoogleNet'') models show robustness to scaling up to 100x the original pixel values. The Caffe implementation of InceptionV1 follows right behind, see text for discussion. 
    }
    \label{fig:app:lucid_changing_scale}
\end{figure}

\begin{figure}
    \includegraphics[width=\columnwidth]{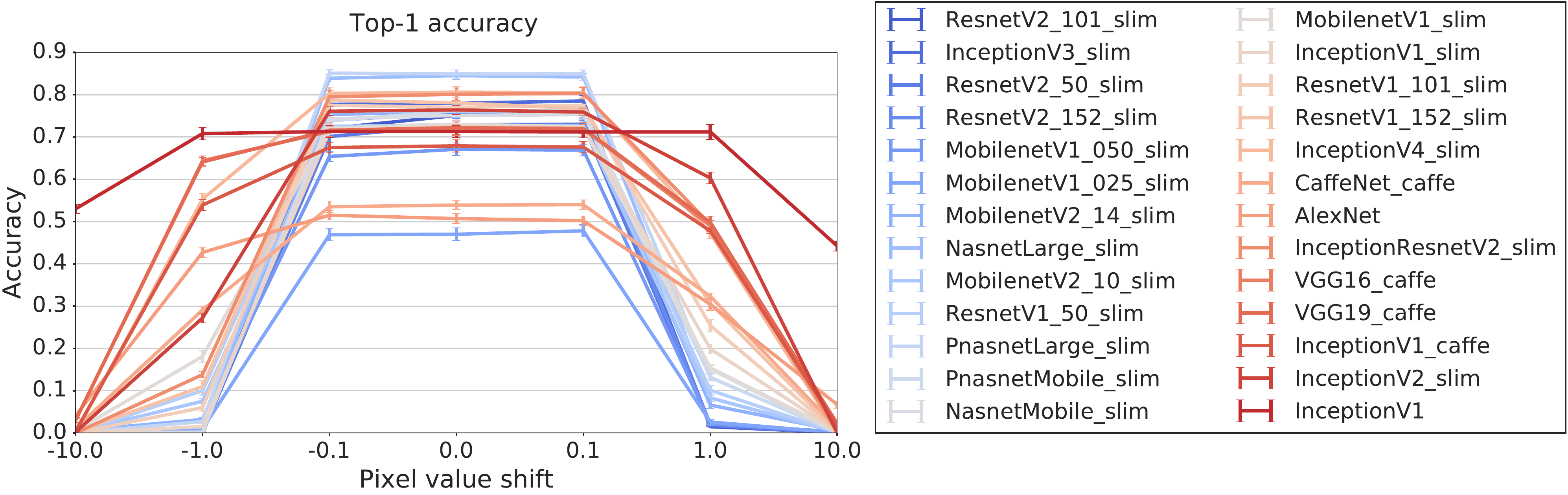}
    \caption{Robustness of modern ImageNet classifiers to the shift in pixel values. The shift is measured relative to the range of pixel values each network was trained on (255 or 1). Colour reflects relative change in performance when shifting pixel values by a full value range (e.g. adding 255).
    }\label{fig:app:lucid_changing_center}
\end{figure}

We study robustness of image classifiers to rescaling (Fig.\,\ref{fig:app:lucid_changing_scale}) and shifting the mean (Fig.\,\ref{fig:app:lucid_changing_center}) of the input values. 
The experiments were performed with all \bn\ parameters fixed, as published \cite{lucid} and according to how they are deployed in the world. Re-evaluating the mean and variance parameters helps to increase the robustness (as observed in our \resnet\ implementation).

Overall, VGG and Inception V1 networks (except for InceptionV1\slim, which was trained with Batch Normalization) appear most robust to our input manipulations. 
So how do these networks differ from the rest? 

First, they do not use \bn. Although many networks trained with \bn\ appear robust when the brightness is decreased, this might be due to the introduction of data augmentation, a technique that is now widely used.  

Second, VGG networks do not deploy any non-linearity beyond rectification. GoogleNet (InceptionV1) uses Local Response Normalization, but with a unique choice of parameters, different from those used in the less rrobust networks: AlexNet, CaffeNet, InceptionV1\caffe networks. Recall, 
\begin{equation}
    LRN(x_{ik}) = x_{ik} / (b + a  (\sum_{j \in (k, k+d)} x_{ij})^2)^\beta
\end{equation}
where $x_{ik}$ is the activity of the $i$-th neuron of the $k$-th convolutional kernel (channel). GoogleNet uses $a=1e-4$, $\beta=0.5$, $b=2$, $d=5$, while the other networks use $a=0.2e-4, \beta=0.75,  b=1, d=2$.

We suspect that GoogleNet's superior robustness to scaling is due to the $\beta$ parameter set ot 2, cancelling out the $(\sum_{k, k+d} x_{ik})^2$ non-linearity for large values of $x_{ik}\gg b/a$. 

Finally, as we have shown all these 6 networks do not rely heavily on bias parameters, we note that scaling the input in a zero-bias system (e.g. the linear classifier $f(\x)=\mathbf{w}\x$), does not affect the ranking of the classes, which is most likely an important factor. 

\subsection{Batch Normalization}\label{sec:app:batch_normalization}
\begin{figure}
    \includegraphics[width=\columnwidth]{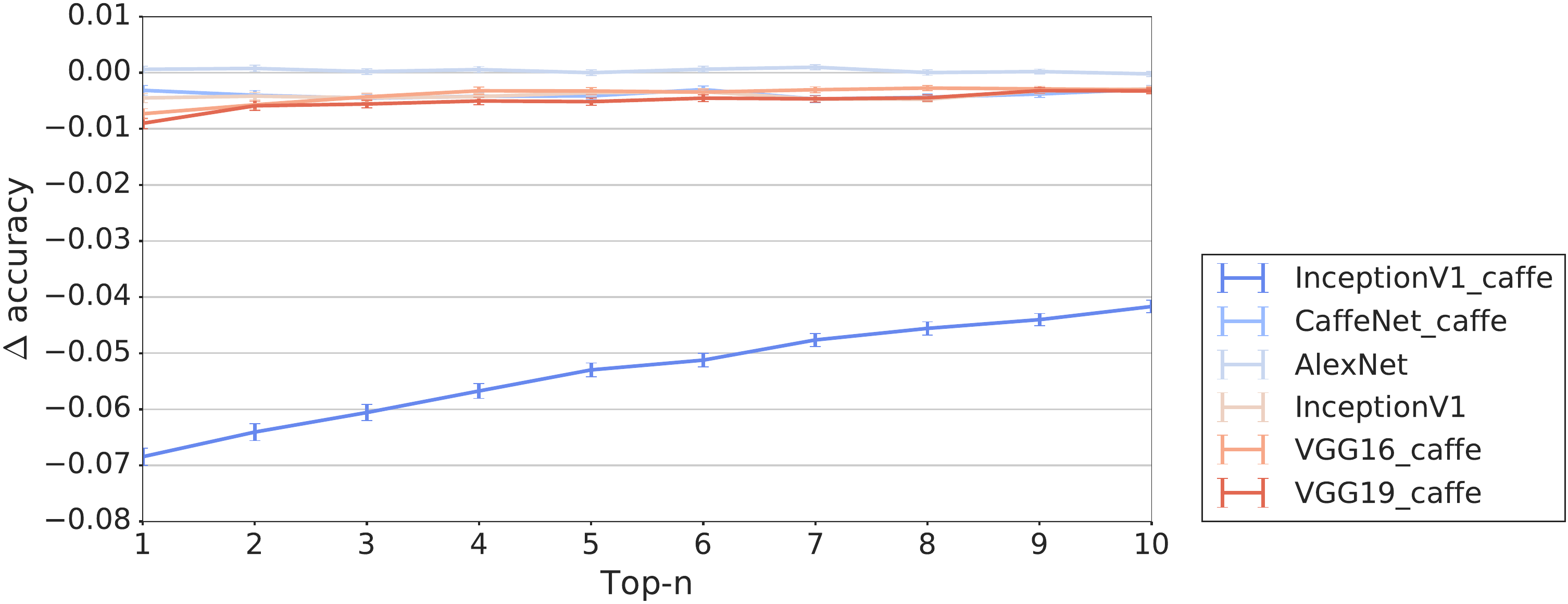}
    \caption{Difference in performance after removing the bias. In the networks robust to bias removal, performance drops by less than 1\% accuracy, except for the Caffe version of InceptionV1. Error bars indicate the mean and standard error over 50000 test images.
    \label{fig:app:accuracy_zero_bias_diff}}
\end{figure}

With the advent of Batch Normalization, the field moved away from using bias as a parameter directly, except for the final layer\footnote{Many of the networks still use both: all 6 ResnetV2 and 2 MobilenetV2 networks and InceptionV2\slim, where it is probably used by mistake, as it appears in a single convolutional layer (``separable convolutions'').}. 
The effective bias stemming from the Batch Normalisation is:
\begin{equation}
\mathrm{\bn}(x_{k}) = \beta - \gamma \frac{\mathrm{mean}(x_k)}{\sqrt{\mathrm{var}(x_k) + \epsilon},}\label{eq:bn}
\end{equation} 
where $\beta$ and $\gamma$ are trainable parameters and $x_k$ is activity of a given neuron (or a convolutional kernel). Activity statistics are collected online, however, at test time, these also act as parameters. 

Of all the ImageNet classifier networks investigated, only  AlexNet, CaffeNet, VGG16/19, InceptionV1 (aka GoogleNet) and InceptionV1\caffe, do not use Batch Normalization. Setting biases to zero in these networks has little effect on the top-1 performance (Fig.\,\ref{fig:app:accuracy_zero_bias_diff}).

\subsection{What do biases do for networks w/o \bn?}
\begin{figure}
    \includegraphics[width=\columnwidth]{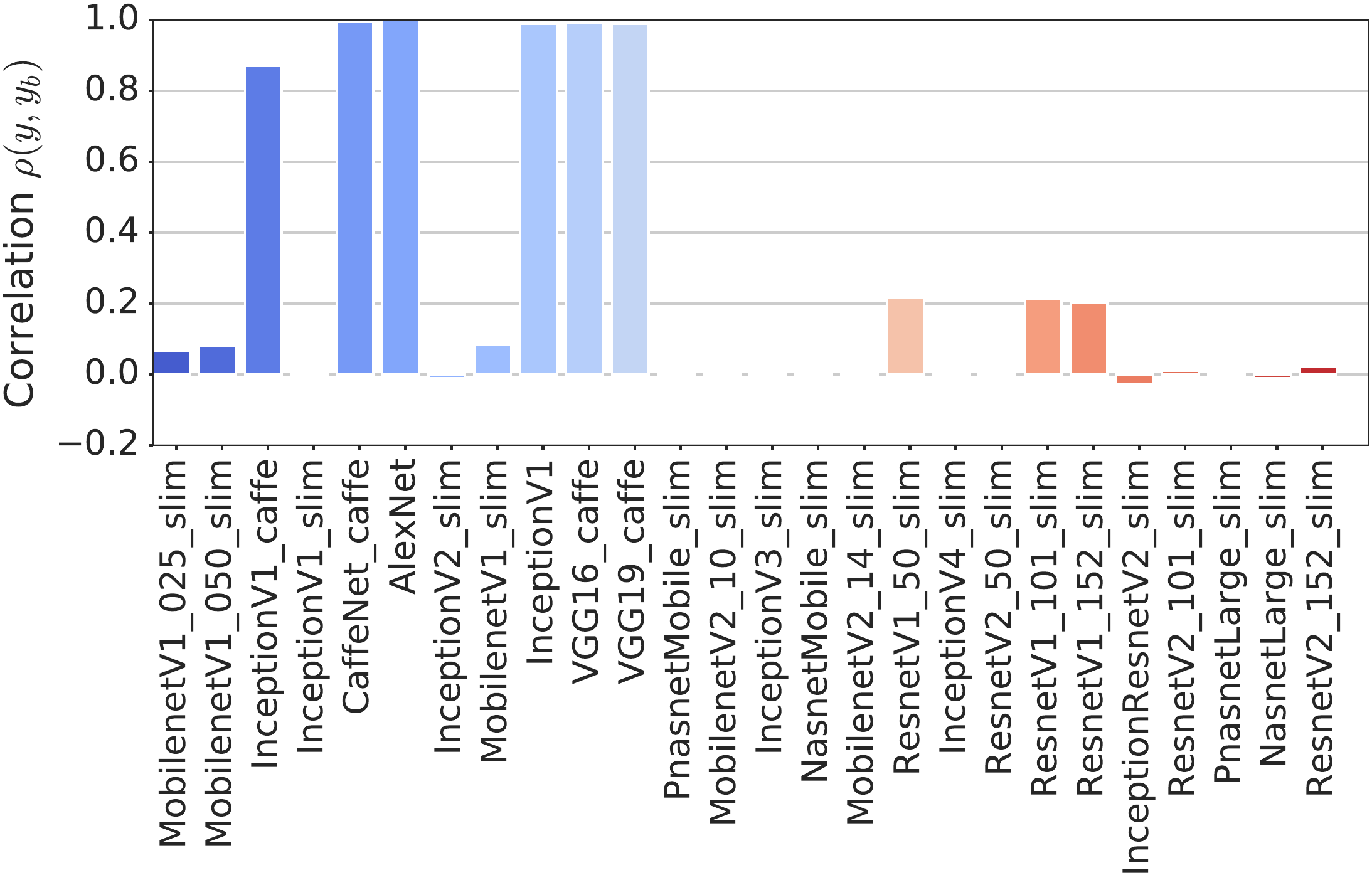}
    \caption{Correlations between outputs of the vanilla networks and the networks with biases removed (mean correlation evaluated over $n=5000$ test images).\label{fig:app:correlation_zb}}
\end{figure}
\begin{figure}
    \includegraphics[width=\columnwidth]{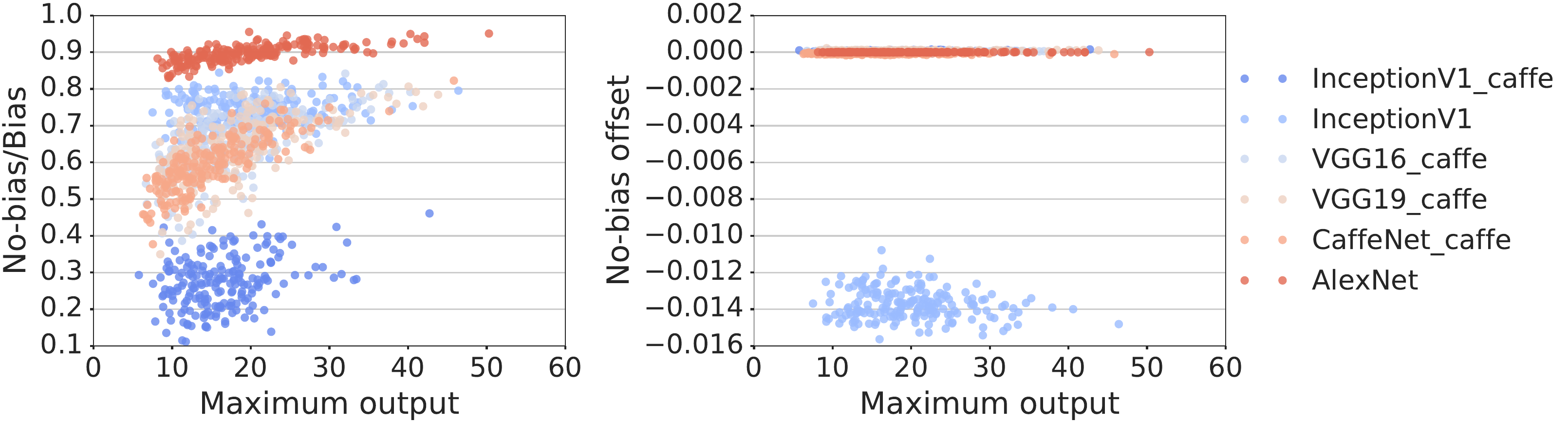}
    \caption{Setting bias parameters to zero increases uncertainty in VGG, AlexNet, CaffeNet and Inception V1 networks. We regress bias-free network output over the vanilla output: $f^{b=0}_c = \alpha f_c+ \beta$ with $c\in [1, 1000]$ (ImageNet classes). Scale coefficients $\alpha$ (left) and offsets $\beta$ (right) are shown for 200 such fits. 
    Offset is negligible compared to the range of output values (even for InceptionV1, $\beta$ is 4 orders of magnitude less than the maximum $f_c$).
    \label{fig:app:acorrelation_zero_bias}
    }
\end{figure}
Does it mean that in the networks not using \bn\ biases play no role in these networks? The answer is no. Although the outputs before and after bias removal are highly correlated (Fig.\, \ref{fig:app:correlation_zb}), they are not equal. In fact, the output after bias removal is very well approximated by a simple multiplication $\yb=\alpha \y$, as shown in Fig.\,\ref{fig:app:acorrelation_zero_bias}. The scaling factor $\alpha$ is always less than 1, meaning the networks become more uncertain (the probability distribution over the outputs, obtained by squashing and re-normalising $f_c$-s, flattens).

\section{Decaying biases}\label{sec:app:bias_decay}
\begin{figure}
    \centering
    \subfigure[\vgg]{
    \includegraphics[width=.99\columnwidth]{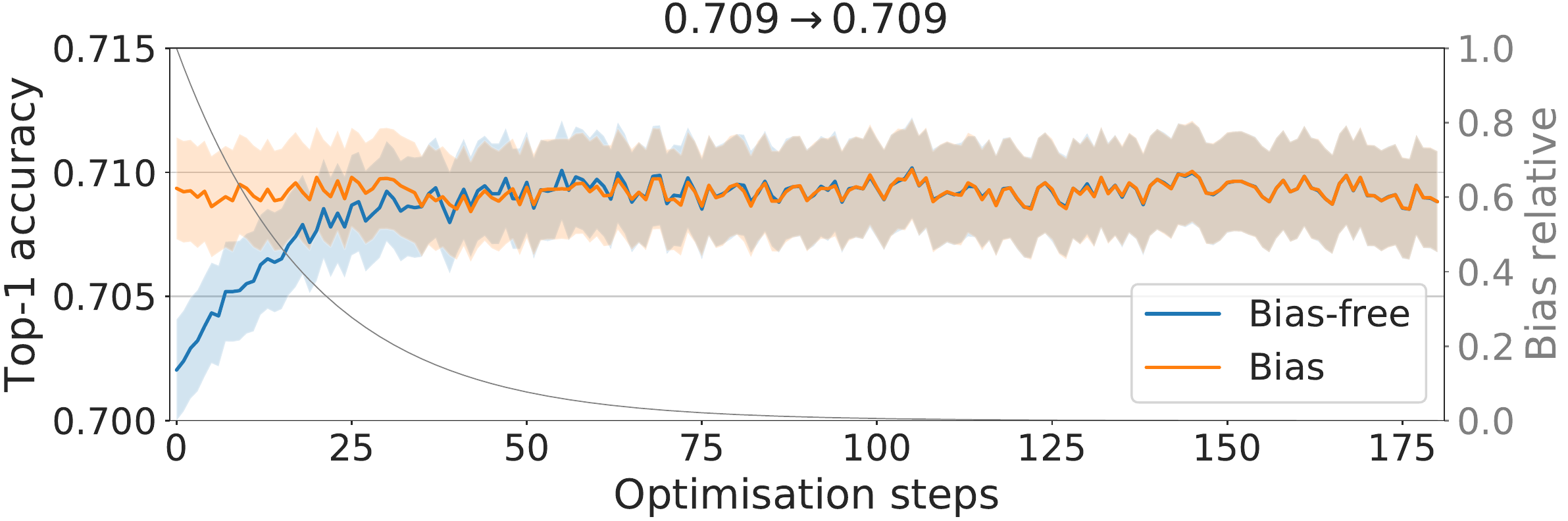}\label{fig:app:decay_vgg}}
    \subfigure[\resnet]{
    \includegraphics[width=.99\columnwidth]{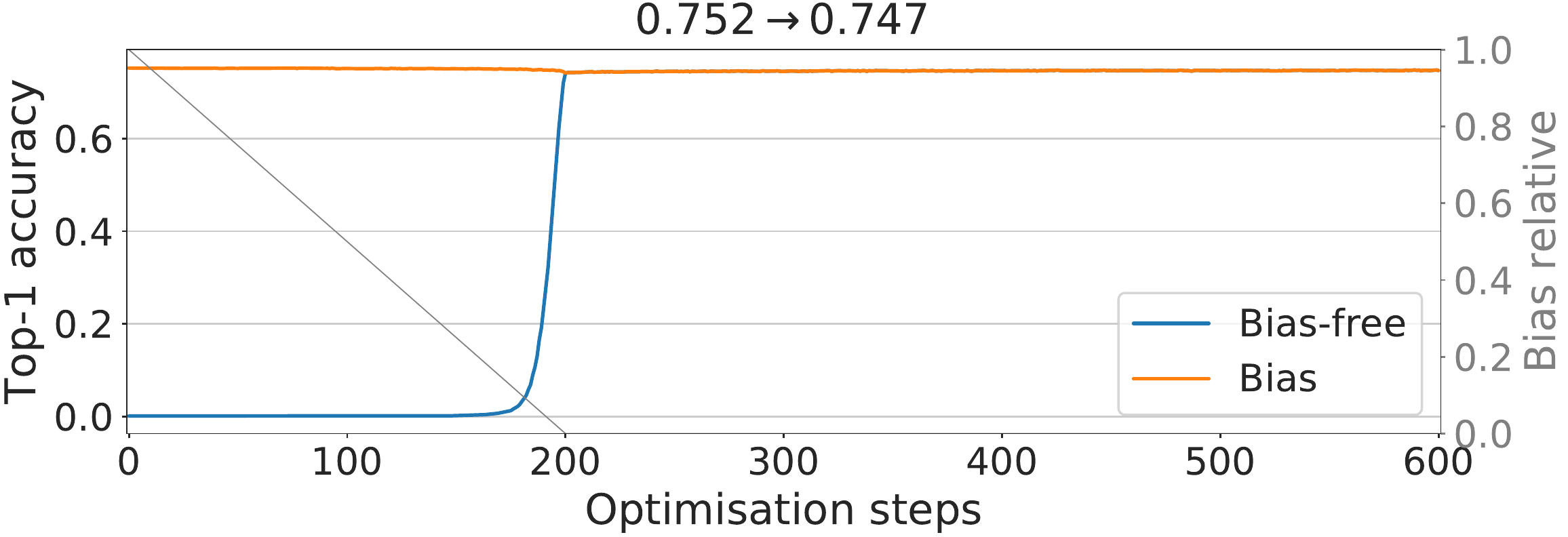}
    \label{fig:app:decay_resnet}}
    \caption{Decaying biases. Decay profile (in grey) and performance of the network (orange), compared to the performance by brute force setting the biases to zero (blue). The network is fine-tuned to keep the output as close to the original network as possible, with temperature parameter $T=100$ \cite{Hinton2015}. Each step shown corresponds to at least 200 training steps (at a given value for biases).}
\end{figure}

\begin{figure}
    \centering
    \includegraphics[width=\columnwidth]{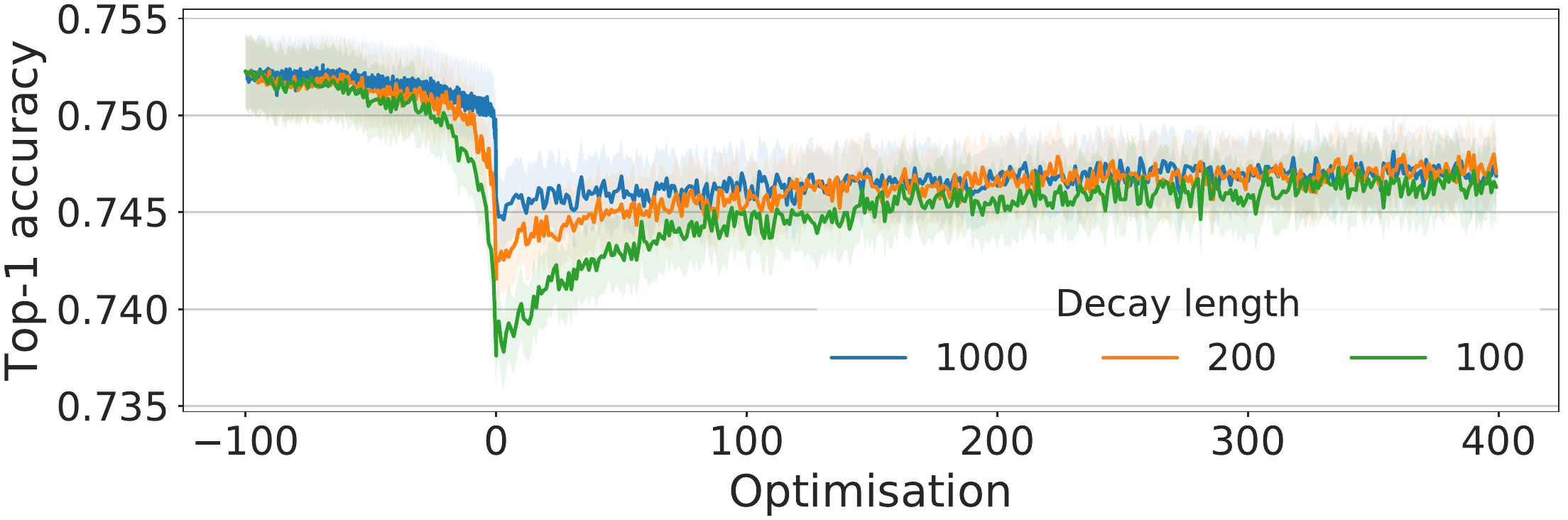}
    \caption{Decaying bias in \resnet\ with T=100
    We tried a different number of steps for decaying the bias (see Fig.\,\ref{fig:app:decay_resnet}). We also tried T=10, T=50 and T=1000, all of which led to a larger drop once the bias was set to zero, and had a harder time recovering from that drop.
    \label{fig:app:decay_resnet_steps}}
\end{figure}

Both in VGG16 and \resnet, we decay all bias parameters from their original values while retraining the networks on the ImageNet dataset. In \resnet, biases include the \bn\ parameters:  $\mathrm{mean}(x_k)$ and $\beta$ (Eq.\,\ref{eq:bn}). We keep the other parameters $\gamma$ and $\mathrm{var}(x_k)$ constant throughout the training. 
We re-scale the values of biases every 200 training steps (batch size = 64). We tried both the exponential decrease (Fig.\,\ref{fig:app:decay_vgg}) and a linear one for ResNet (Fig.\,\ref{fig:app:decay_resnet}). In ResNet, we added extra training with biases set to zero. We also tested different speeds of the bias decay in this network and settled on 200 decay steps (Fig.\,\ref{fig:app:decay_resnet_steps}). 

The temperature parameter for the cross-entropy cost function between the original and the fine-tuned network that worked best in both \vgg\ and \resnet\ was $T=100$ \cite{Hinton2015}. We used the Adam optimiser with the learning factor set to $5 e^{-6}$. We did not run hyper parameter sweeps, as obtaining a perfect bias-free version of ResNet was not central for this research.

\section{Does completeness matter?}\label{sec:app:saliency_biasfree}
\begin{figure}
    \centering
    \subfigure[\vgg]{
    \includegraphics[width=.9\columnwidth]{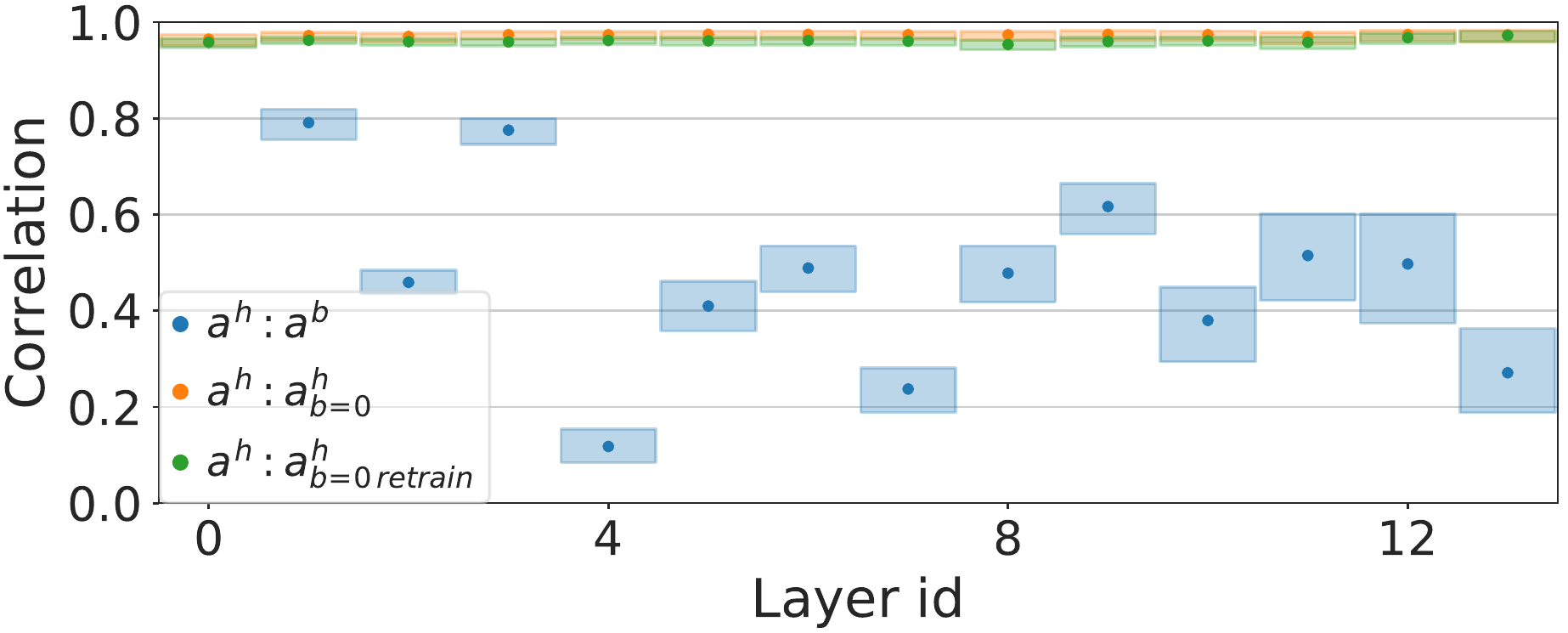}\label{fig:app:correlations_vgg}
    }
    \subfigure[\resnet]{
    \includegraphics[width=.9\columnwidth]{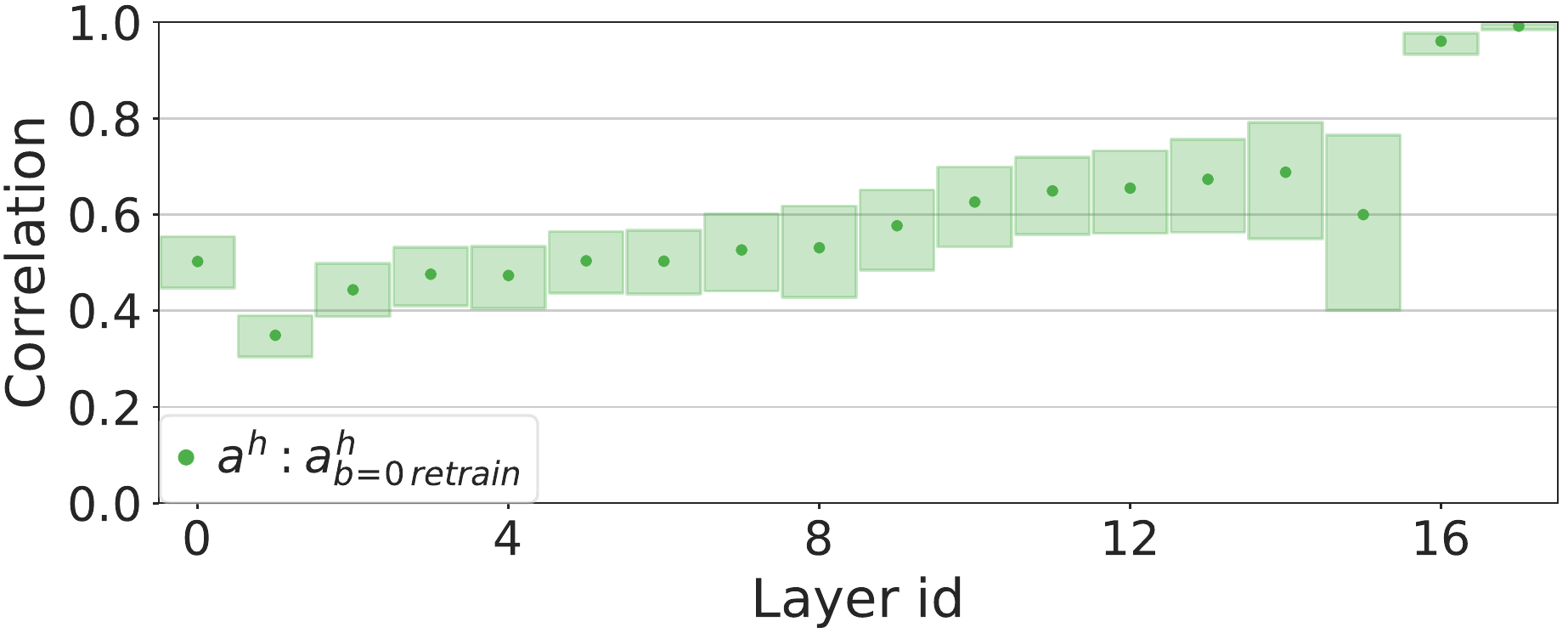}\label{fig:app:correlations_resnet}
    }
    \caption{Spatial correlations between attributions in \vgg.
    Pearson's correlation coefficient between activity $\ahl$ and bias $a^{b,l}_c$ attributions in the vanilla \vgg\ (blue), and between $\ahl$ from the \vgg\ and our zero-bias versions. Depicted statistics are: median, 10th and 90th percentile over 100 classes x 20 samples.
    For every layer, $a^{b,l}_c$  are positively correlated with the $\ahl$, but not for every layer $\rho$ is high. Activity attributions change very little when bias parameters are set to zero (orange). Re-training the network only slightly changes $\ahl$  across the network (green).}
    \label{fig:app:correlations}
\end{figure}

In Figure \ref{fig:app:correlations}, we quantify the similarity between the original ``incomplete'' $\ahl$ from zero-bias networks, with biases set to zero either by brute force (orange) or by decay (green). 
In \vgg, the similarity is high for all layers, with Pearson's correlation coefficient $\rho>0.95$ (Fig.\,\ref{fig:app:correlations_vgg}). In \resnet, the similarity is high only in the top two layers, it drops as low as $\rho\sim 0.3$ in the first convolutional layer (Fig.\,\ref{fig:app:correlations_resnet}).


We proceed to evaluate whether the visual explanations we derive from $\ahl$ improve with completeness.

\begin{figure}
    \centering
    \includegraphics[width=\columnwidth]{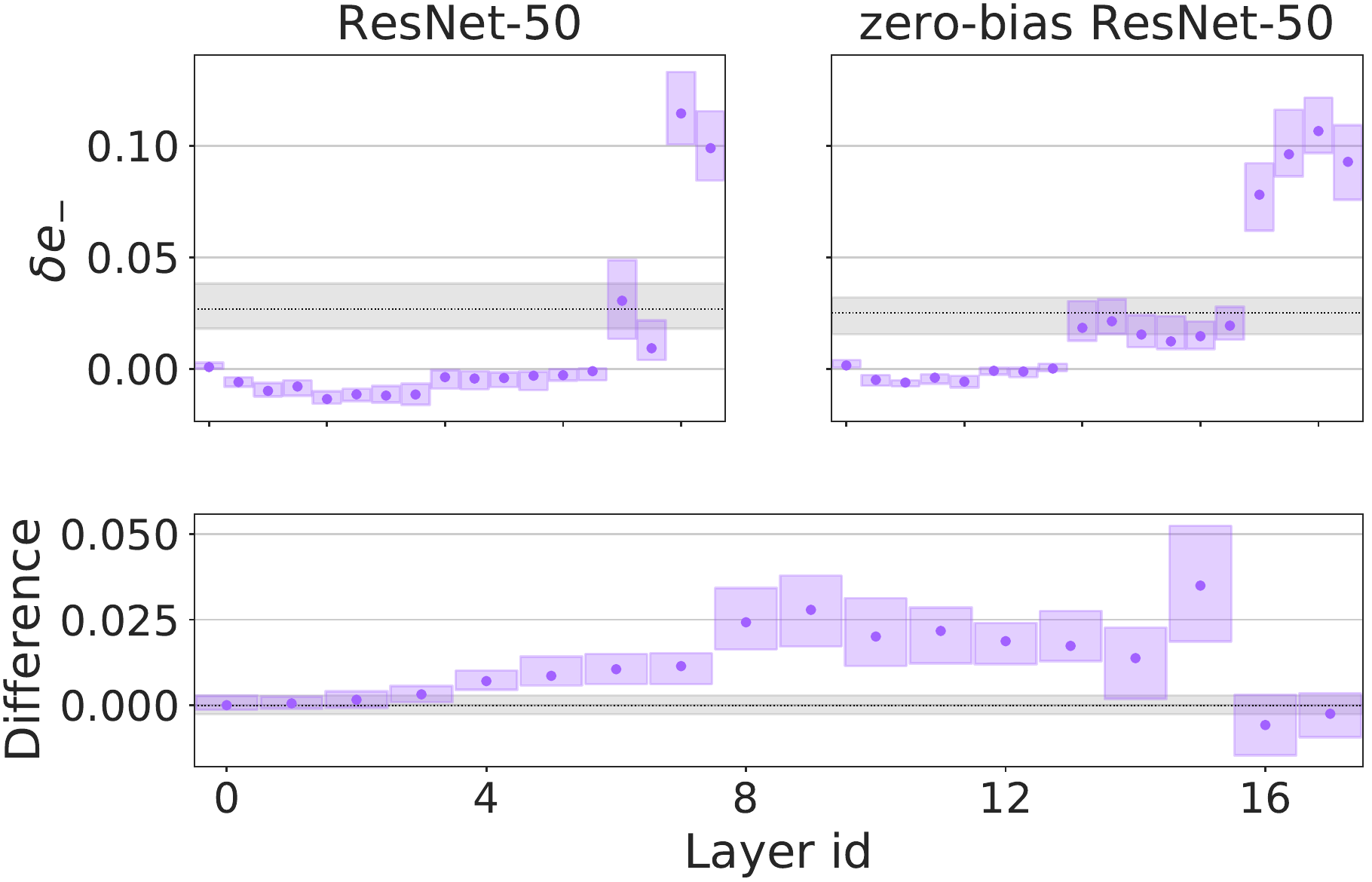}
    \caption{zero-bias networks yield superior signed explanations. When using activity attributions from a zero-bias \resnet\ (top right), the explanations appear to be more faithful in estimating the sign of evidence from each pixel than in the vanilla \resnet\ (top left). The difference (bottom) is smaller for the unsigned attributions (blue) and it diminishes in the top two layers, as well as for the input attribution ($0^{\text{th}}$-layer). Gradient-based method ($s=\sum_{RGB}(\abs(g_x))$) is shown for reference in grey.
    Boxes represent 25-th to 75-th quantiles.}
    \label{fig:app:perturbation_bias_free_attributions}
\end{figure}

In order to use $\ahl$ as attribution maps, we match their size to that of the input image. For upsampling, we use bilinear interpolation, as in \citet{SelvarajuEtAl2016, Srinivas2019}:
\begin{equation}
\s^{h,l}\gets\mathrm{bilinearUpsample}(\ahl)\label{eq:attribution_to_saliency}.
\end{equation}
The linear scaling is more prone to artefacts and yields higher variance, see  \S\,\ref{sec:app:bilinear}.

In Figure \ref{fig:app:perturbation_bias_free_attributions}, we show perturbation results for the original $\ahl$ attributions (top left) and those obtained from a zero-bias version of the network (top right), with their difference shown in the bottom plot. 
The metric shown is $\delta \ee$ (Eq.(\ref{eq:delta_e})); the higher values denote more discriminative explanations.
As a baseline, we report the quality of the standard gradients approach (grey).  Any visible differences are highly significant as the medians are collected over 100x50 samples. The colored boxes represent 25$^{\text{th}}$--75$^{\text{th}}$ percentiles.

Clearly, the completeness underlying attributions from the zero-bias \resnet\  has a positive effect on the quality of attribution maps. 
The improvement increases with depth, except for the two top layers  (significance ranges from $p<10^{-9}$ in layer 2 to $p<10^{-100}$ in layer 9; Wilcoxon signed rank test). Unfortunately, the top two layers are the ones affording the best visual explanations in \resnet. That means decaying bias parameters did not enhance upon the best attribution maps.

\begin{figure}
    \centering
    \includegraphics[width=.99\columnwidth]{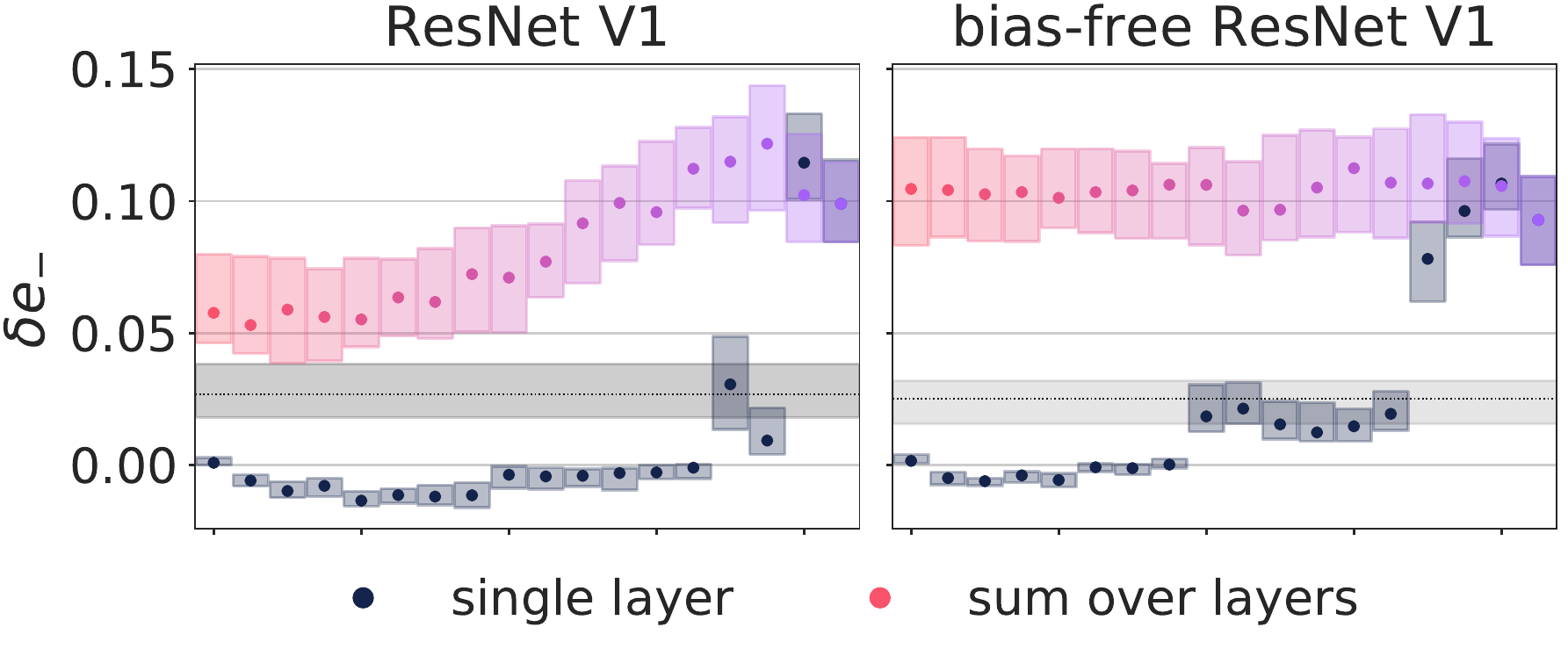}\\
    \includegraphics[width=.99\columnwidth]{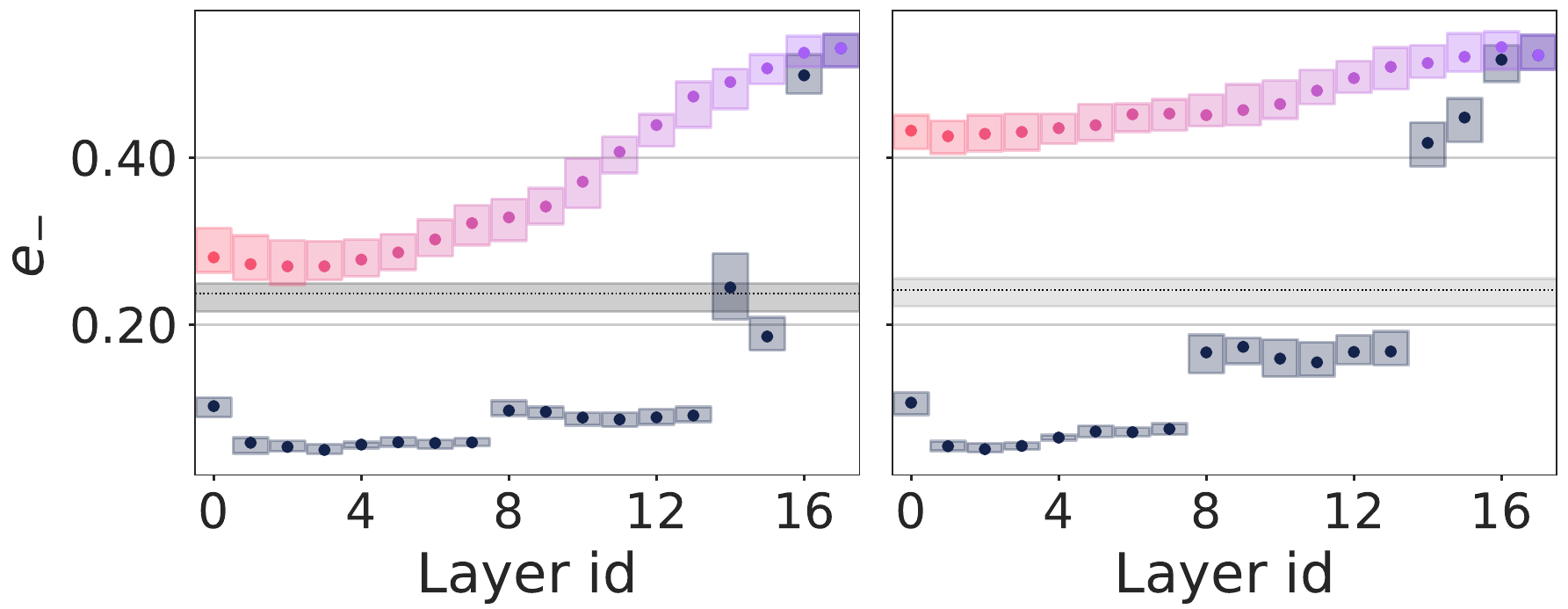}
    \caption{Averaging attributions over layers.
    Single layer explanations are shown in black. Aggregate explanations are formed by averaging single-layer attribution maps from top to bottom (purple to red). In zero-bias \resnet, aggregating layers has less detrimental effect than in the vanilla \resnet.}
    \label{fig:app:perturbation_averaging_over_layers}
\end{figure}
Akin to results presented in Fig.\,\ref{fig:perturbation_averaging_over_layers}, we tested attribution maps aggregated over layers. We see that aggregating  attribution maps from the zero-bias \resnet\ prevents the strong deterioration of the quality observed with the vanilla \resnet, as measured with $\delta\ee$ (Fig.\,\ref{fig:app:perturbation_averaging_over_layers}, top). A large improvement in the general quality of the visual explanations can be also seen in $\ee$  (Fig.\,\ref{fig:app:perturbation_averaging_over_layers}, bottom).

\subsection{Saliency maps}
\begin{figure}
    \centering
    \includegraphics[width=\columnwidth]{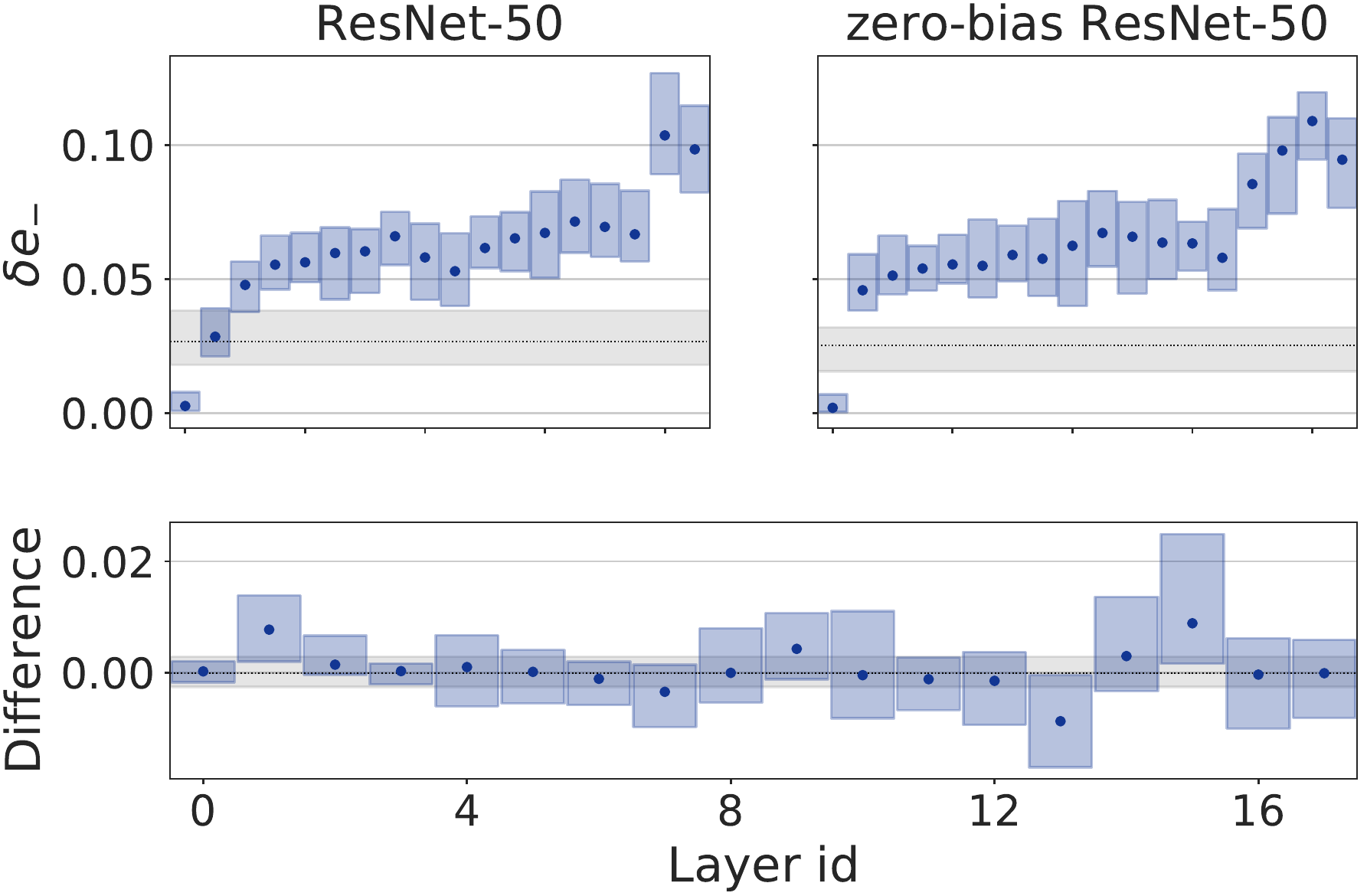}
    \caption{Zero-bias networks yield superior signed explanations. When using activity attributions from a zero-bias \resnet\ (top right), the explanations appear to be more faithful in estimating the sign of evidence from each pixel than in the vanilla \resnet\ (top left). The difference (bottom) is smaller for the unsigned attributions (blue) and it diminishes in the top two layers, as well as for the input attribution ($0^{\text{th}}$-layer). 
    }
    \label{fig:app:perturbation_bias_free}
\end{figure}

We perform pixel removal perturbations by using saliency maps, rather than attribution maps, i.e. working with $\Psi(\abs(\ahl))$, rather than $\Psi(\ahl)$. Figure \ref{fig:app:perturbation_bias_free_attributions} presents the results as in Fig.\,\ref{fig:app:perturbation_bias_free}. 

The best single layer explanations available in \resnet\ come from the top two layers of the network. Choosing least important pixels according to $\s^{h,L}$, one can remove an extra 10\% of the image, as compared to what can be removed using random explanations. This largely improves over the gradient-based approach, which helps to remove only about an extra 2\% of the image signal. 

Finally, setting bias parameters to zero does not improve these best explanations. The median of pair-wise differences between $\delta\ee$ from the original and bias-free \resnet\ lies down on the zero line (Supplementary \ref{fig:app:perturbation_bias_free}, bottom). The same is true for the input  
attributions ($l=0$), where most saliency methods focused so far. While some improvement can be seen for layers $l=1$ (where zero-bias \resnet\ to improve beyond the gradients approach, $p<10^{23}$), and $l=15$ ($p<10^{13}$)), the overall effect is not very strong.

\section{Choose bilinear over linear up-sampling}\label{sec:app:bilinear}
\begin{figure}
    \centering
    \includegraphics[width=\columnwidth]{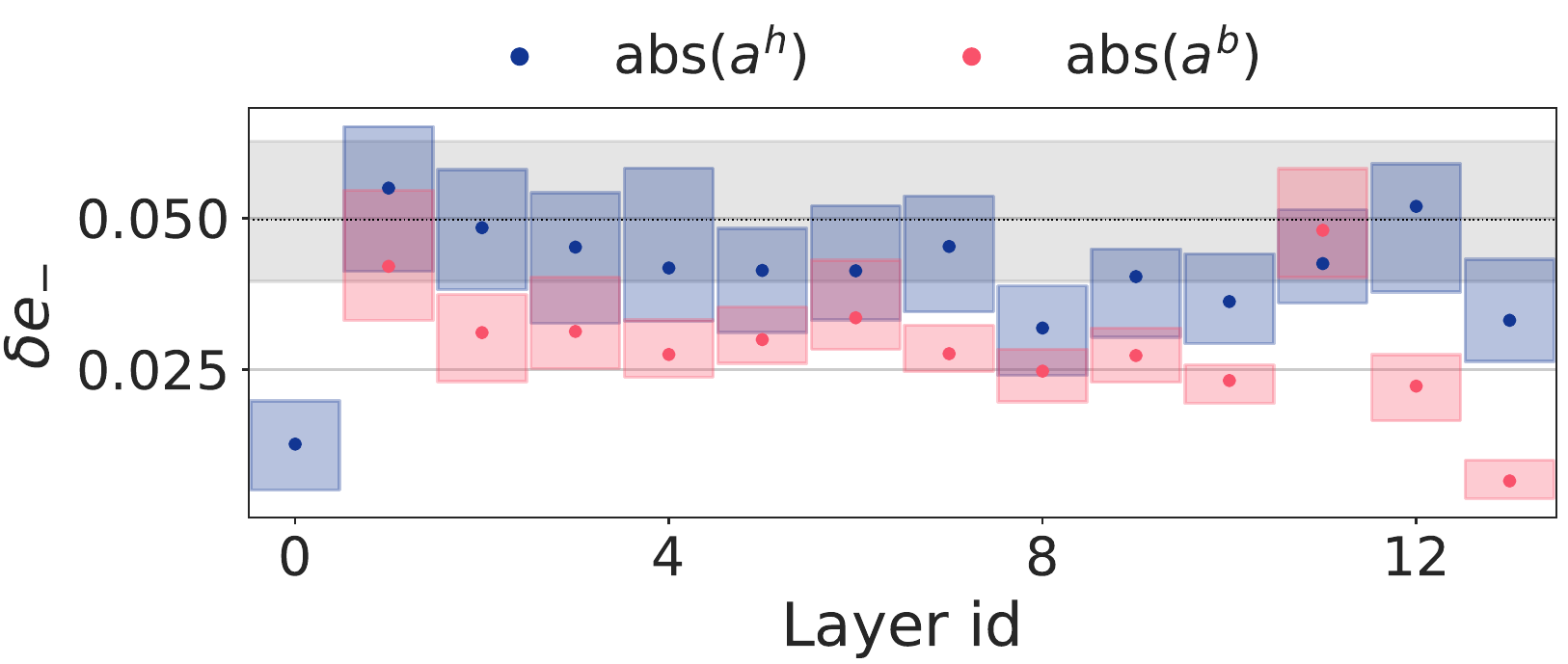}
    \caption{Activity attributions are more salient than bias-related attributions. Same as in Fig.\,\ref{fig:perturbation_vgg_ab} from the main text, but with a linear, rather than a bilinear, resize operation.}
    \label{fig:app:perturbation_vgg_ab_linear_resize}
\end{figure}
We provide a qualitative assessment of the impact of the choice of up-sampling operation, when generating saliency maps (Eq.\,(\ref{eq:attribution_to_saliency})). In Fig.\,\ref{fig:app:perturbation_vgg_ab_linear_resize}, we present results as in Fig.\,\ref{fig:perturbation_vgg_ab} in the main text, but with a bilinear up-sampling replaced by linear one. The effect size of $\delta\ee$ is 2$\times$ smaller, all methods appear to be less discriminative than the simple gradient approach. However, the difference between $\ahl$- and $\abl$-derived explanations is still highly significant (except for layer 11).

\begin{figure}
    \centering
    \includegraphics[width=\columnwidth]{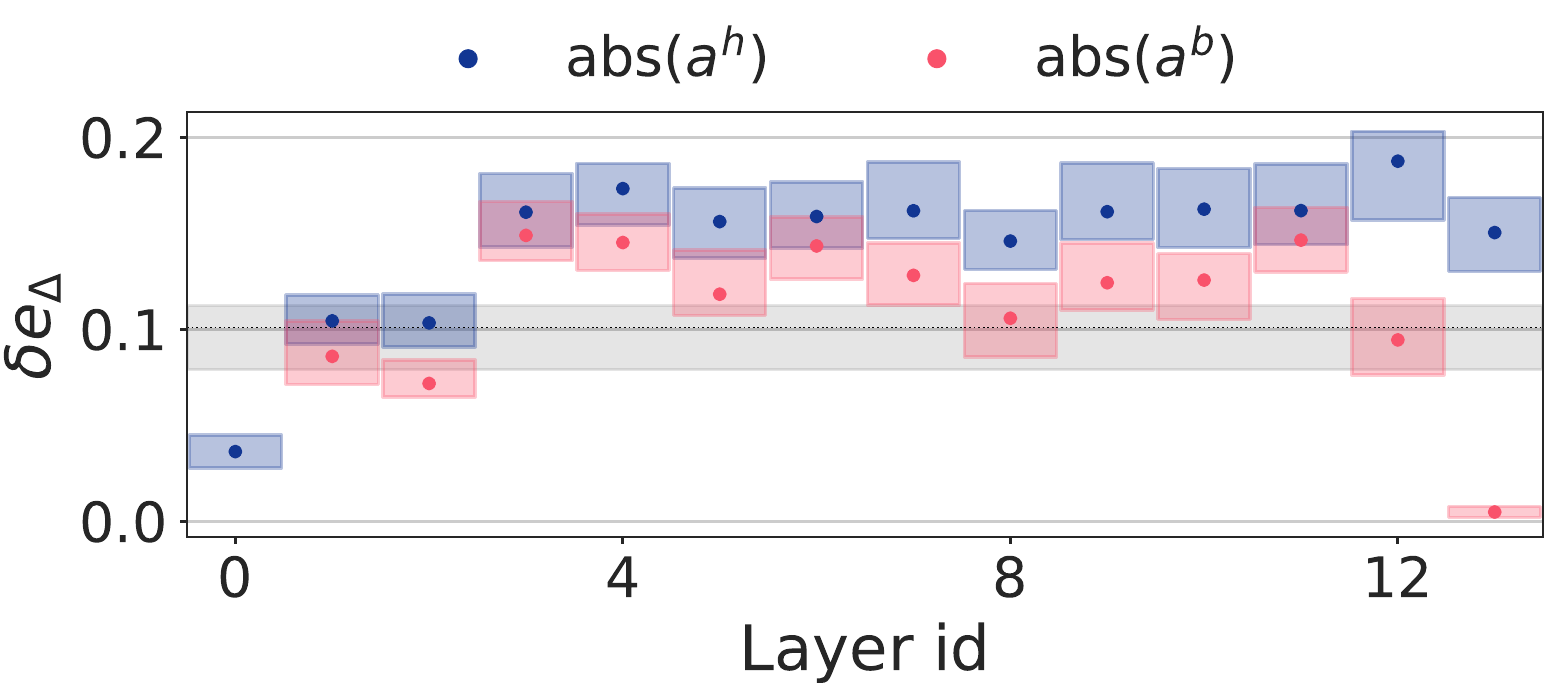}
    \caption{Activity attributions are more salient than bias-related attributions. Same as in Fig.\,\ref{fig:perturbation_vgg_ab}, but with $\e_\Delta$ as a metric.}
    \label{fig:app:perturbation_vgg_ab_pxrem_ds}
\end{figure}
Similarly, as noted before (\S\ref{sec:app:perturbations}), $\delta\e_\Delta$ is a better choice of the metric.
It increases statistical significance of our tests by up to 30 orders of magnitude. We decided against using it in the main text purely for the sake of simplicity of description. Qualitatively, all our results  were unchanged by the choice of metric.

\section{Evaluation of perturbation methods}\label{sec:app:perturbations}
\subsection{The three perturbation metrics}
\begin{figure}
    \centering
    \includegraphics[width=.99\columnwidth]{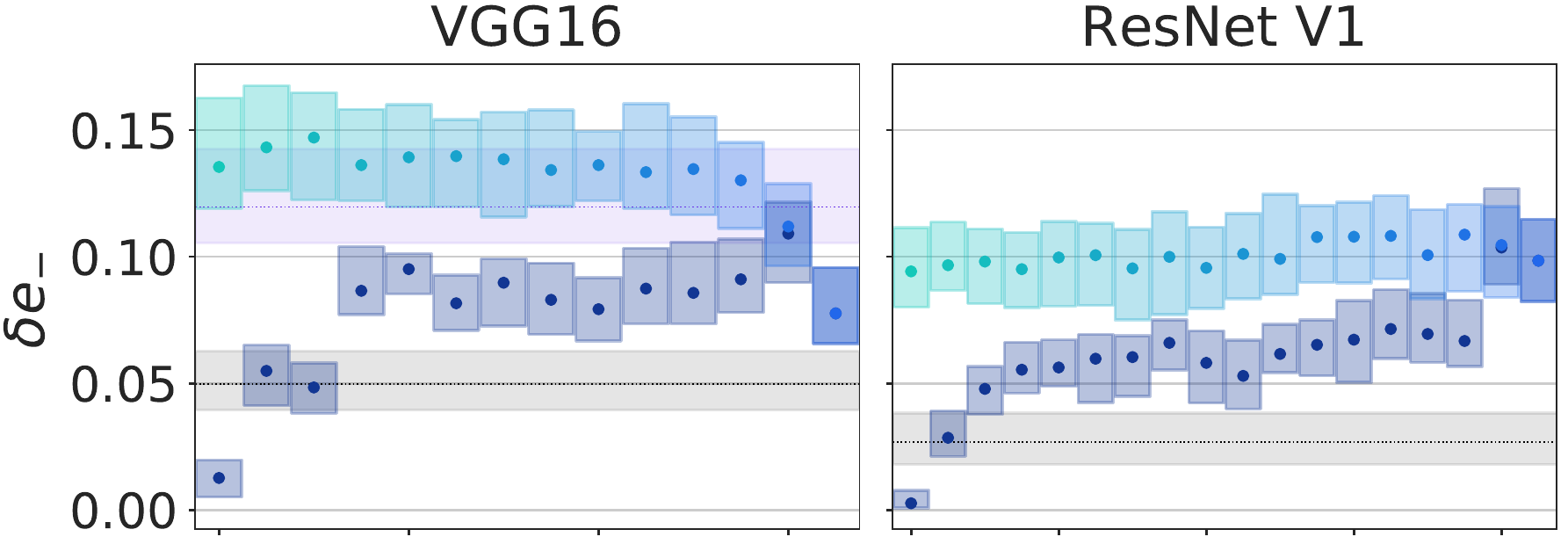}\\
    \includegraphics[width=.99\columnwidth]{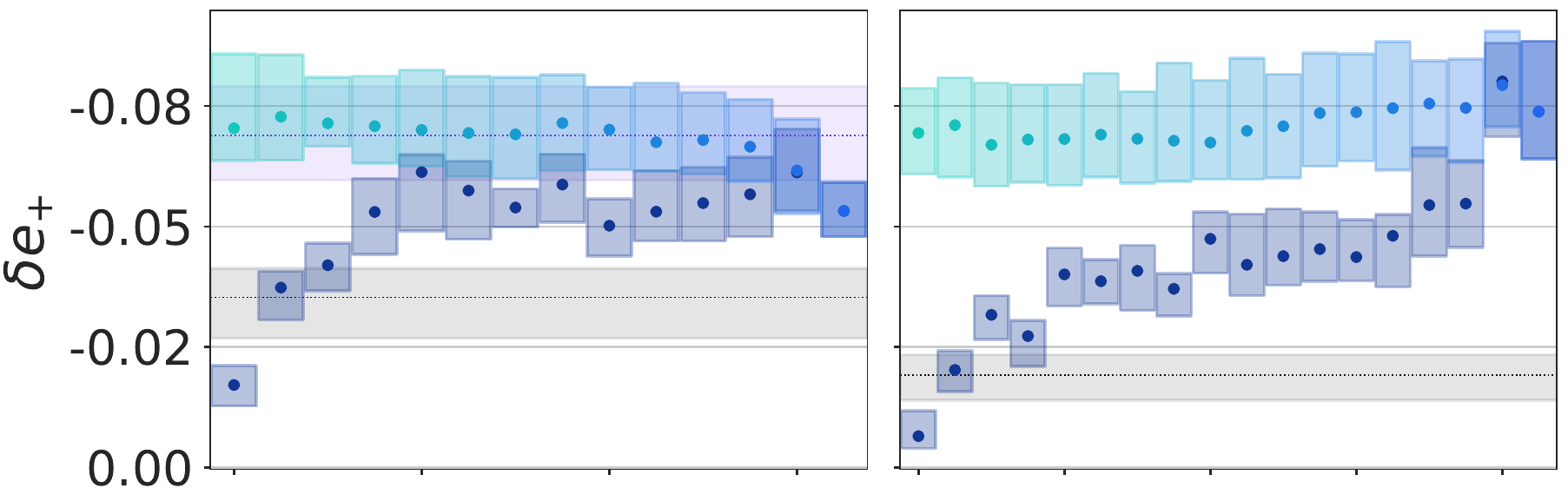}\\
    \includegraphics[width=.99\columnwidth]{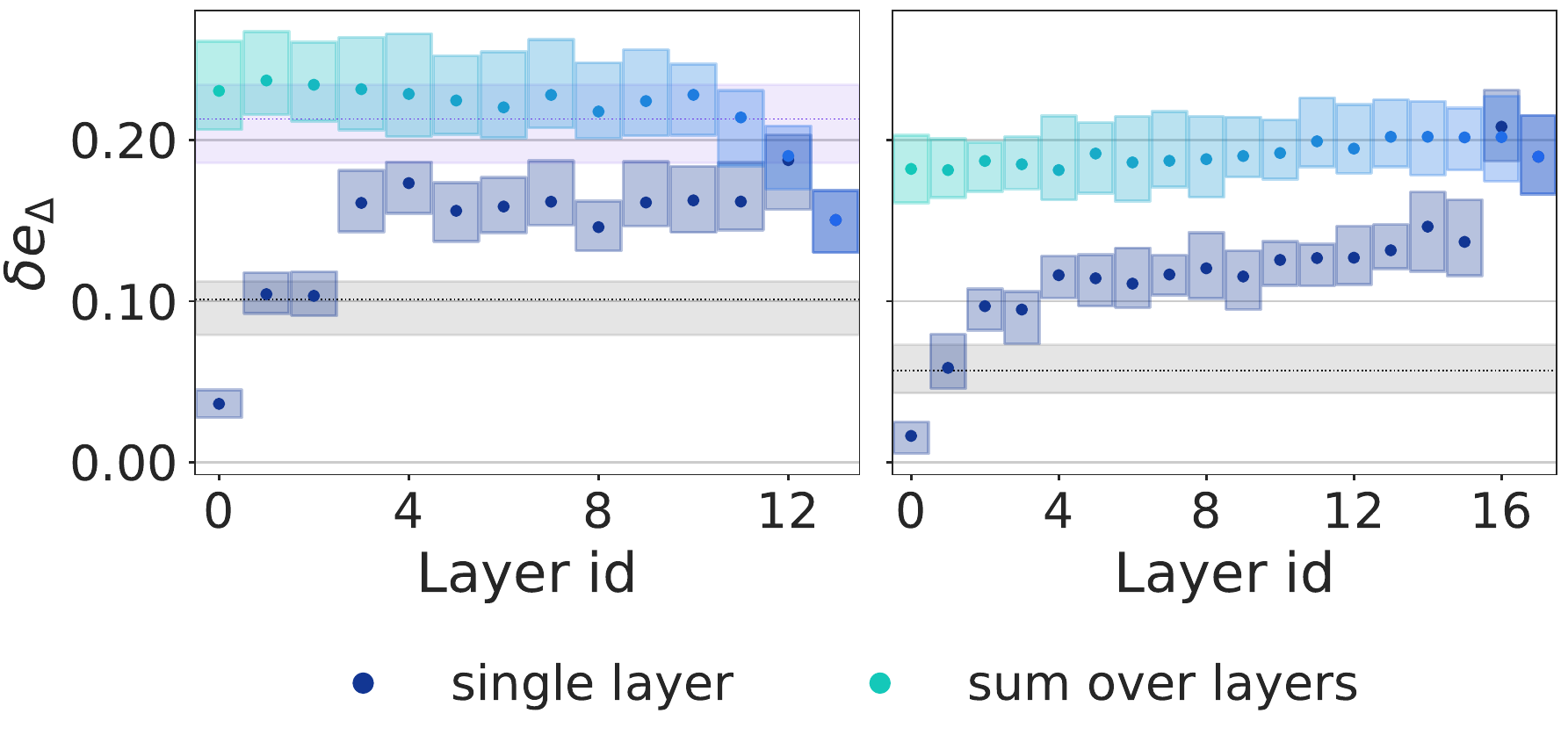}
    \caption{Our main result on the discriminatory power of saliency maps, as measured with $\delta\ee$, $\delta\e_+$ and $\delta\e_\Delta$. Top panel is the same as in Fig.\,\ref{fig:perturbation_averaging_over_layers}, top. Starting pixel removal from the least important pixels (top) and most  important pixels (second row) yields qualitatively similar results (note the y-axis in $\delta\e_+$ reversed to ease the comparison). The sum of the two effects (bottom) increases statistical significance.}
    \label{fig:app:pxrem}
\end{figure}
\begin{figure}
    \centering
    \includegraphics[width=.99\columnwidth]{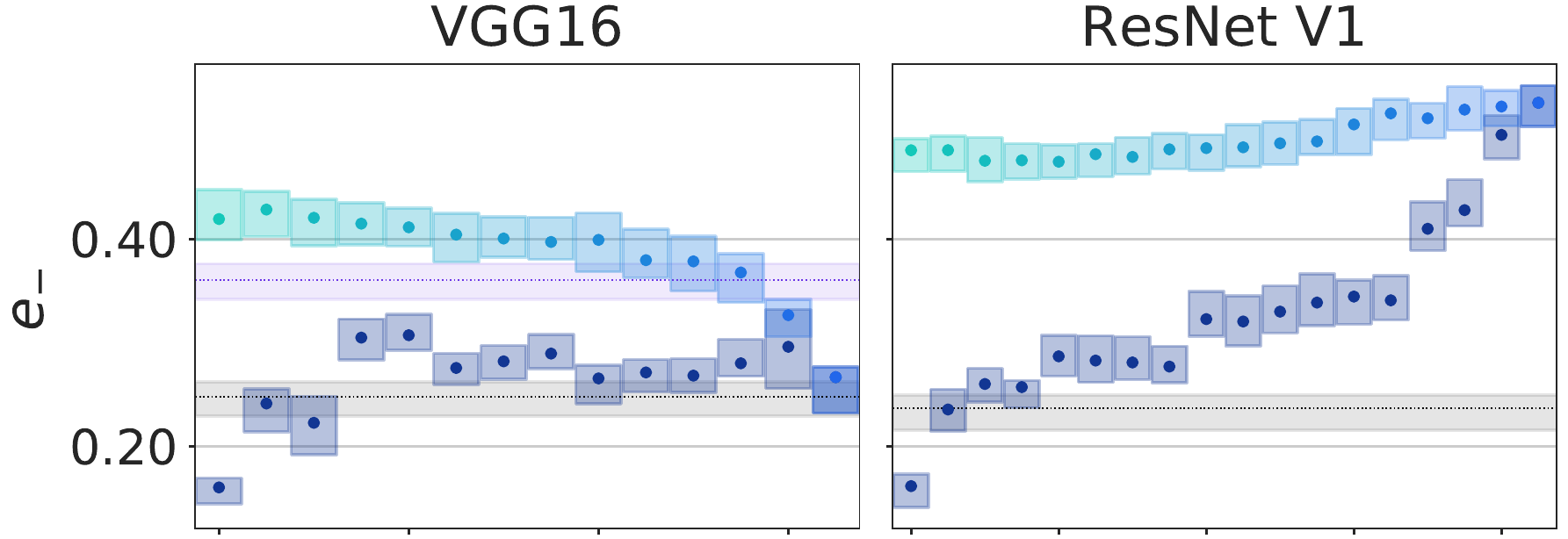}\\
    \includegraphics[width=.99\columnwidth]{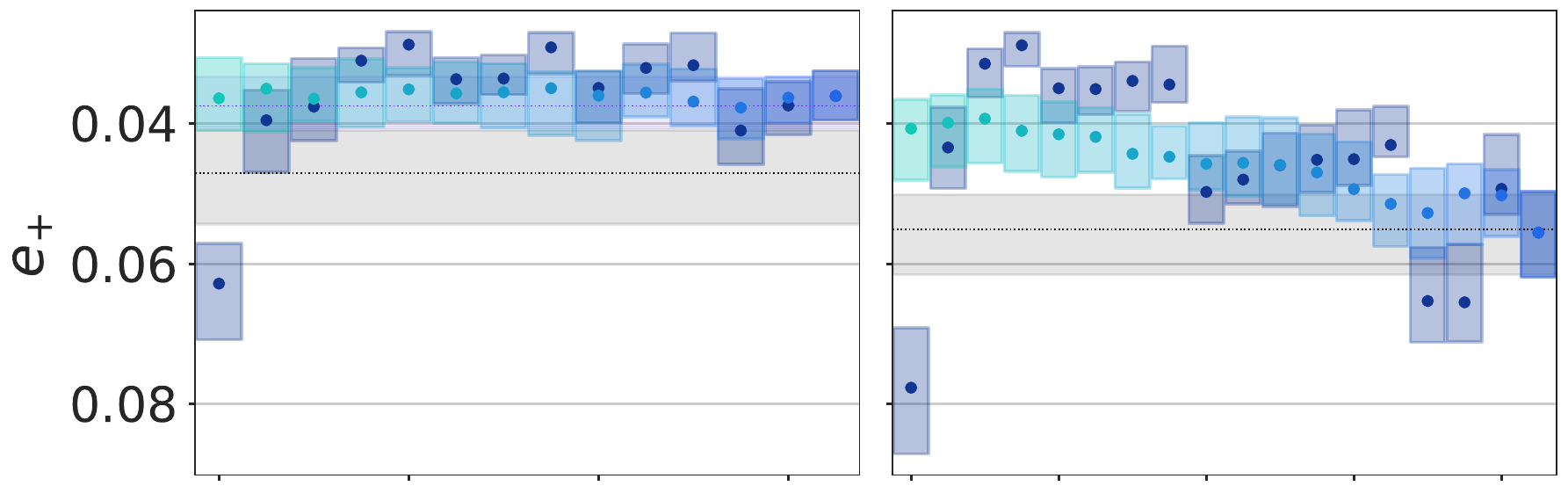}\\
    \includegraphics[width=.99\columnwidth]{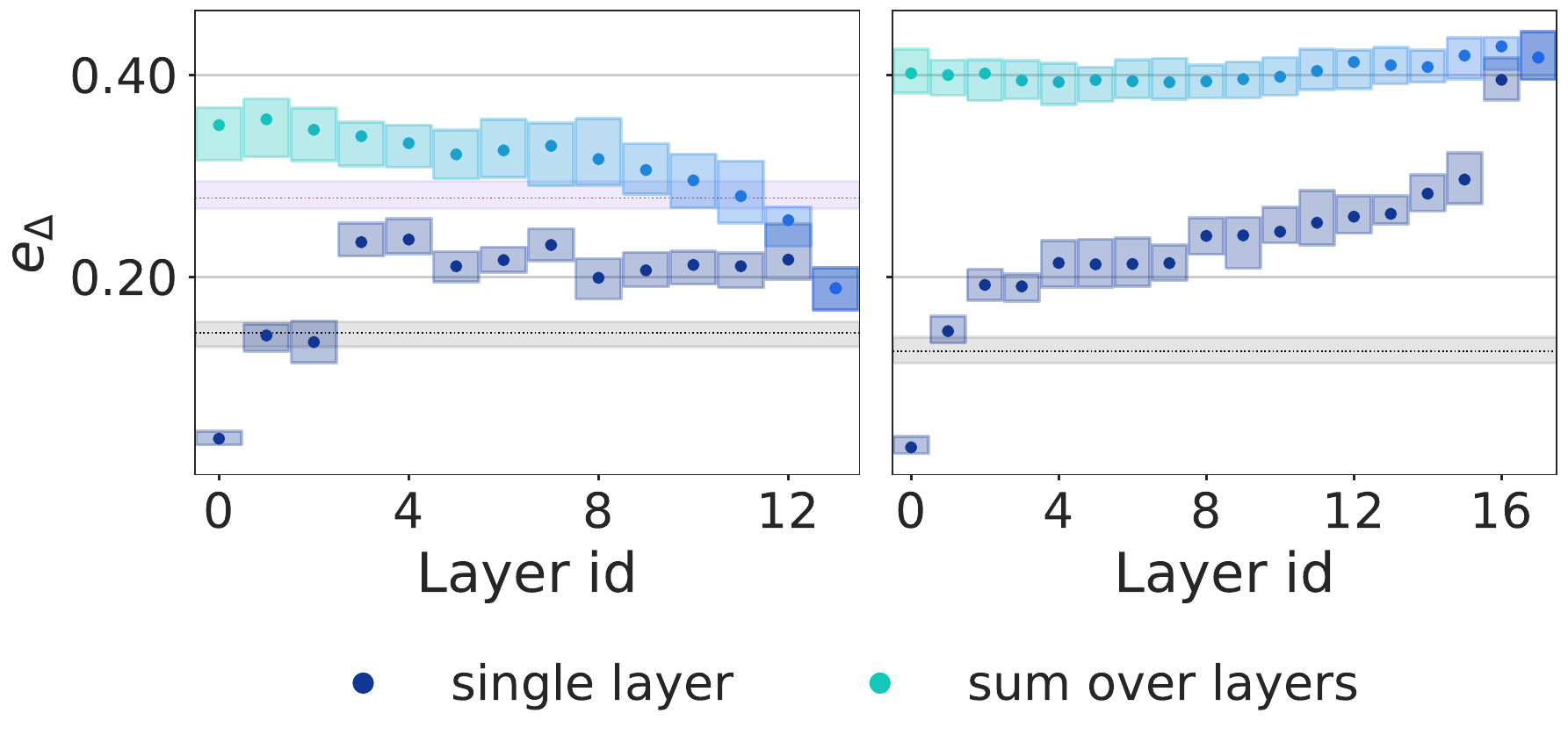}
    \caption{Our main result on the overall perturbation effect, as measured with $\ee$, $\e_+$ and $\e_\Delta$. Top panel is the same as in Fig.\,\ref{fig:perturbation_averaging_over_layers}, bottom.  Starting pixel removal from the least important pixels (top) yields qualitatively different results from the  $-$ approach. In \vgg, all methods appear to produce saliency maps of similar quality. This shows the importance of taking noisy perturbations into account (Fig.\,\ref{fig:app:pxrem}).}
    \label{fig:app:pxrem_signal}
\end{figure}

In Figures \ref{fig:app:pxrem} and \ref{fig:app:pxrem_signal}, we present our final results, as in Fig.\,\ref{fig:perturbation_averaging_over_layers}, comparing the metric used in the main text ($\ee$, perturbing least salient pixels) with  
$\e_+$ (perturbing most salient pixels), and $\e_\Delta$ (the difference between the two effects). Note that we reverse the y-axis when depicting $\e_+$, to ease the comparison with the other metrics, so that in all figures higher means better.
As in the main text, $\delta$ denotes a comparison to the perturbation using noisy explanations (\S\,\ref{sec:evaluating_expectations}).

We notice a good agreement between the $\delta\ee$ and $\delta\e_+$ metrics (Fig.\,\ref{fig:app:pxrem}). Consequently, the sum of the two effects (bottom) yields a stronger effect size and higher statistical significance. 

Note, that the effect size of $\delta\ee$  is about 2$\times$ stronger than the $\delta\e_+$. For example, using ``All layers'' approach (``Layer id''=1), one can move 15\% more signal using a valid saliency map as compared to using a noisy saliency map. In contrast, when targeting the most important pixels, only 8\% less signal will be removed before the classification changes, as compared to a noisy perturbation. 
%


Figure \ref{fig:app:pxrem_signal} illustrates how important is the comparison with the noise. When measuring the effects of perturbations directly, the results vary widely between $\ee$ and $\e_+$.

Most strikingly, when measured with $\e_+$, all saliency methods appear similar in effect, including FullGrad (purple) and gradients (grey) approach: Removing 4\% of the important signal ($\e_+$) causes a change in classification, no matter which saliency map is used. Moreover, the ``single layer'' saliency maps show a specific pattern with local maxima at layers 4, 7 and 12 in \vgg\ and 3, 7, 13 in \resnet. 

Removing unimportant pixels yields a much stronger signal (20\%--40\% in $\ee$, rather than $\sim$4\% in $\e_+$), which dominates $\e_\Delta$.
While $\ee$ and $\e_\Delta$ appear qualitatively more similar to $\delta\ee$ (top panels in Figs \ref{fig:app:pxrem} and \ref{fig:app:pxrem_signal}), there is still a clear difference in the details of these metrics.

Only by comparing with noise can we reconcile the results of $+$ and $-$ perturbation approaches. We now turn to these results directly, in order to understand the role of accidental perturbations and artefacts.

\subsection{Comparison with noise is critical}\label{subsec:app:perturbations_noise}
\begin{figure}
    \centering
    \includegraphics[width=.99\columnwidth]{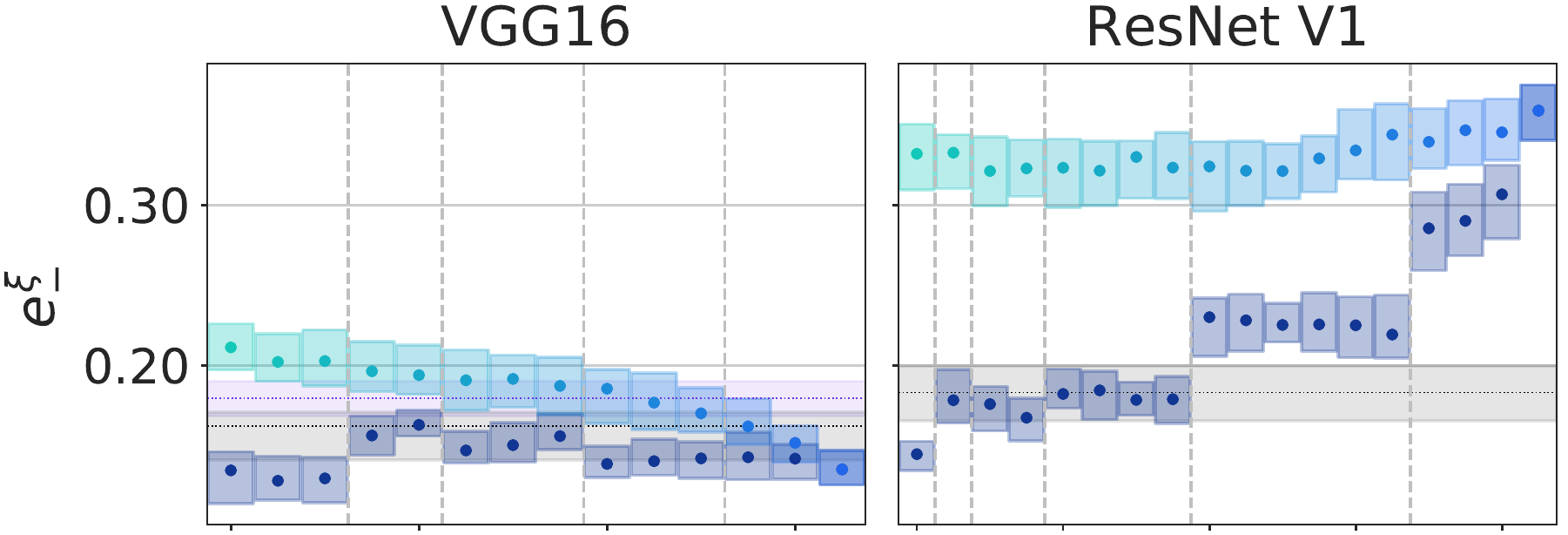}\\
    \includegraphics[width=.99\columnwidth]{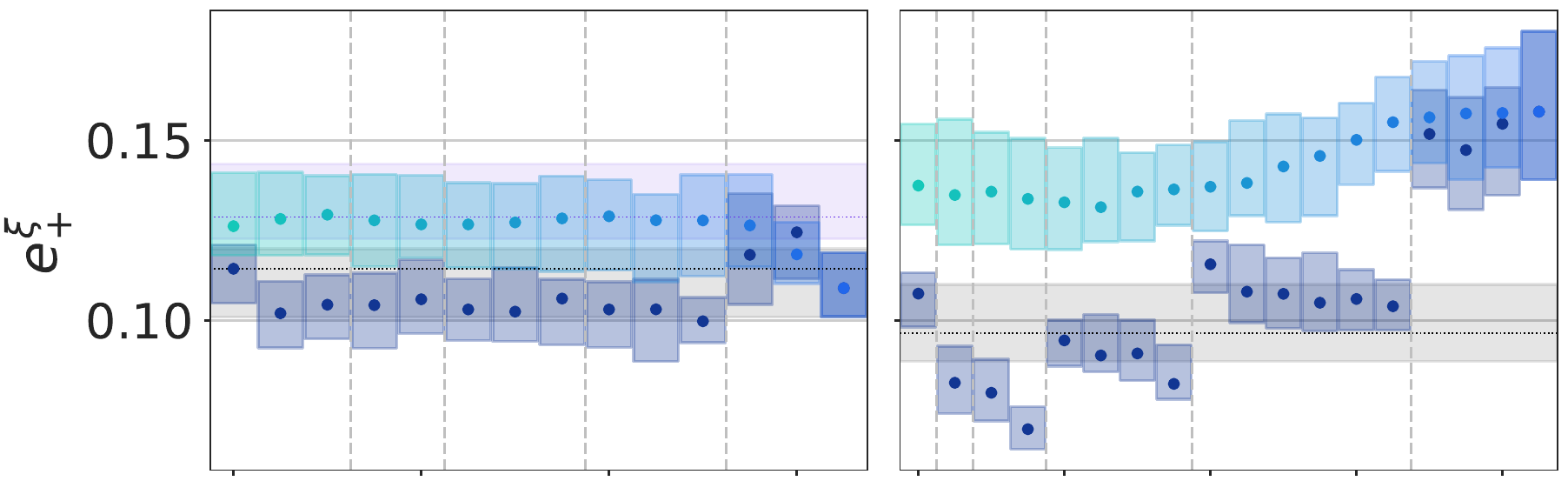}\\
    \includegraphics[width=.99\columnwidth]{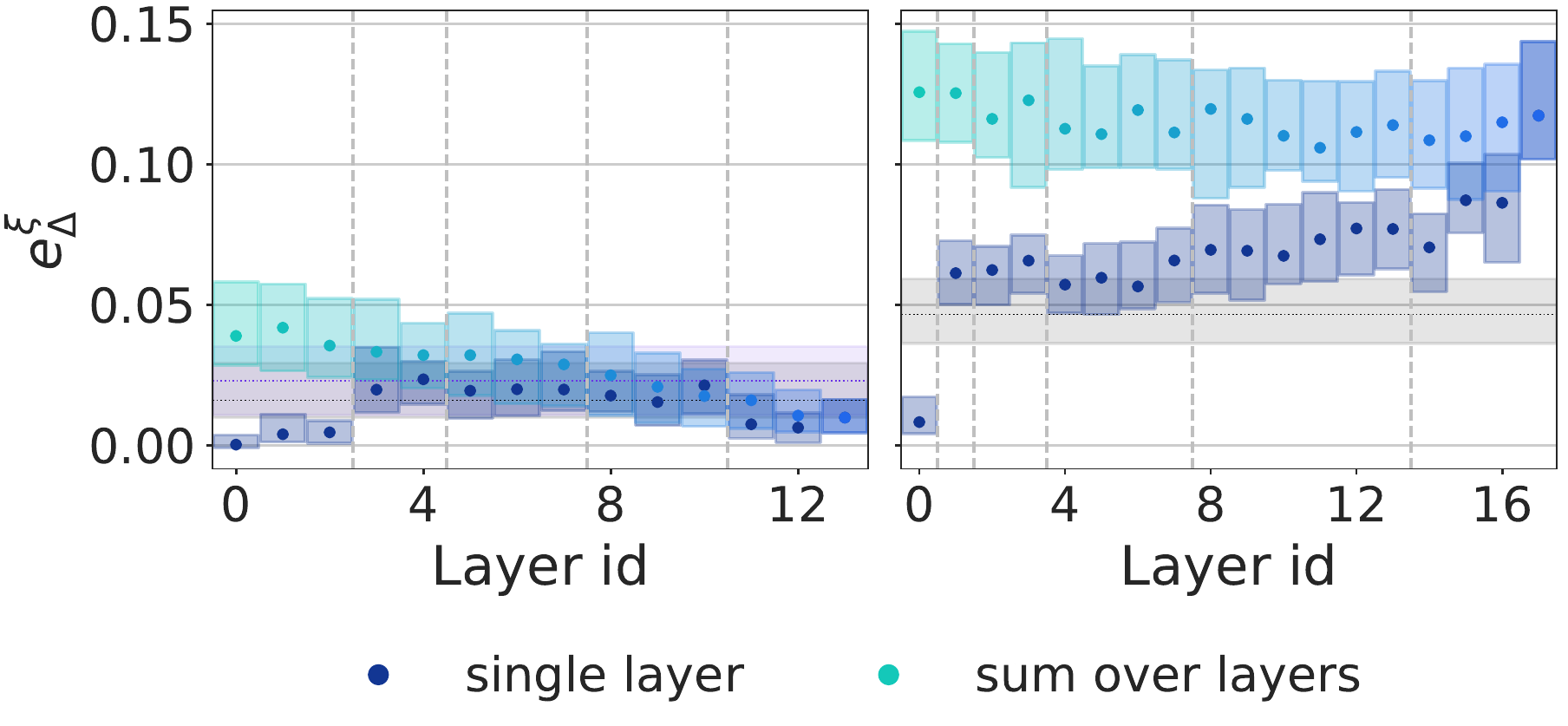}
    \caption{The effect of perturbations with noise. 
    Both $\e^\xi_-$ and  $\e^\xi_+$ summarise how perturbative are the underlying saliency maps for the network. For example, smoother patterns are less likely to change the image decision, which is why the ``sum over layers'' yield higher values of $\e^\xi$ than the ``single layer'' saliency maps.
    Note, the y-axis in the $\e^\xi_+$ plot is not reversed, to ease the comparison (higher values mean less artefactual). The vertical lines indicate halving the convolutional layer size.}
    \label{fig:app:pxrem_noise}
\end{figure}

In Figure \ref{fig:app:pxrem_noise}, we show the $\ee$, $\e_+$, $\e_\Delta$ statistics using the shuffled explanations $\xi$ as the saliency maps. Metrics $\e^\xi_-$ and $\e^\xi_+$ quantify the impact of perturbations abstracted from the image content. They estimate how ``artefactual'' are the patterns of saliency maps extracted with a given method. Their difference, $\e^\xi_\Delta$, should be zero if the artefactual impact was identical in $-$ and $+$ perturbation approaches. Only in that case, could we omit the comparison with noise and use $\e_\Delta$ alone as a valid metric for evaluating visual explanations. 
Unfortunately, for both \vgg\ and \resnet, $\e^\xi_\Delta$ is not only non-zero, but structured, making the comparison with the noise a crucial part of the evaluation.

Let us analyse the outcome of the noisy perturbations in detail. 
First of all, in both \vgg\ and \resnet, aggregating saliency maps over layers (``sum over layers'') yields higher values of  $\e^\xi_-$ and $\e^\xi_+$ (Fig.\,\ref{fig:app:pxrem_noise}, cyan to light blue). This is because aggregating layers top-down, i.e. taking the lowest resolution $\a^{h,L}$ into account, makes the saliency map smoother. In contrast, ``single layer'' saliency maps (Fig.\,\ref{fig:app:pxrem_noise}, dark blue) exhibit higher spatial frequencies and are more likely to perturb the network (decreasing $\e^\xi$).

Single layer saliency maps are different in the two networks. In \vgg, they all seem to have a similar impact on the network---allowing $\sim$10\% of the image signal to be removed without changing the classification. In \resnet, there is a large difference between the maps from layers 1--3; 4--7; 8--13 and 14--17. These steps correspond to the different size of the convolutional layers (112x112 for layer 1; 56x56 for 2--3; 28x28 for 4--7; 14x14 for 8--13; and 7x7 for top layers), marked with vertical lines in Fig.\,\ref{fig:app:pxrem_noise}.

Why do explanations deteriorate with every change of layer size in \resnet, but not in \vgg? One factor could be the overall sizes of layers (in \vgg: 224x224 for layers 1--2; 112x122 for 3--4; 56x56 for 5--7; 28x28 for 8--10 and 14x14 for top layers). However, we suspect that the drop in quality is rather due to the lack of pooling, as we explain below.

In \resnet, there are clear local minima for every layer just before the down-sampling operation, except for layer 1. Similar minima are missing from \vgg. In fact, the down-sampling seems to have a positive effect on the preceding layers, e.g. 4 and 7 (especially clear in Fig.\,\ref{fig:app:pxrem_noise}, top). The difference lies in the architecture: \vgg\ uses max-pooling before every change in layer size. In \resnet, max-pooling is applied only after the first layer. Otherwise, the activity is directly down-sampled by setting the stride=2 in the last convolution of the given block. Down-sampling without pooling creates high-frequency artefacts in $\ahl$, and the related saliency maps.

Note, that the impact of these down-sampling artefacts is similar for $-$ and $+$ approaches to perturbation, because they do not show in the  $\e_\Delta$ (Fig.\,\ref{fig:app:pxrem_noise}, bottom). Thus, in general, when possible, we would recommend to choose $\e_\Delta$ over $\ee$ metric, as it has the potential to lessen the impact of artefacts.

The rapid deterioration of visual explanations for lower layers in \resnet\ is likely the reason why aggregating explanations did not improve the quality of the visual explanations in this network (as compared to \vgg).


\end{document}